\documentclass{article}

\PassOptionsToPackage{numbers, compress}{natbib}

\usepackage[preprint]{neurips_2024}

\usepackage[T1]{fontenc}    
\usepackage[utf8]{inputenc} 

\usepackage{amsmath,amsfonts,bm}

\usepackage{tcolorbox}

\usepackage{xcolor,amsmath}
\usepackage[linesnumbered,ruled,vlined]{algorithm2e}
\DontPrintSemicolon

\usepackage{xcolor}
\definecolor{mygreen}{HTML}{3cb44b}
\usepackage{varwidth}

\SetKwComment{Comment}{\color{green!50!black}\# }{}

\SetKwProg{Function}{def}{:}{}

\SetKwProg{For}{for}{:}{}
\SetKwProg{If}{if}{:}{}
\newcommand{\VarSty}[1]{\textnormal{\ttfamily\color{blue!90!black}#1}\unskip}

\def\eqref#1{equation~\ref{#1}}

\def\1{\bm{1}}

\DeclareMathAlphabet{\mathsfit}{\encodingdefault}{\sfdefault}{m}{sl}
\SetMathAlphabet{\mathsfit}{bold}{\encodingdefault}{\sfdefault}{bx}{n}

\usepackage{hyperref}       
\usepackage{url}            
\usepackage{booktabs}       
\usepackage{amsfonts}       
\usepackage{nicefrac}       
\usepackage{microtype}      
\usepackage{xcolor}         

\usepackage{wrapfig}

\usepackage{graphicx}
\usepackage{xcolor,colortbl}

\usepackage{minitoc}
\usepackage{multirow}
\usepackage{comment}
\usepackage{enumitem}

\definecolor{Gray}{gray}{0.93}

\definecolor{uclagold}{rgb}{1.0, 0.7, 0.0}
\definecolor{airforceblue}{rgb}{0.36, 0.54, 0.66}
\definecolor{rosegold}{rgb}{0.72, 0.43, 0.47}
\definecolor{pastelbrown}{rgb}{0.51, 0.41, 0.33}
\definecolor{isabelline}{rgb}{0.96, 0.94, 0.93}

\definecolor{macaroniandcheese}{rgb}{0.98, 0.89, 0.83}
\definecolor{wildblueyonder}{rgb}{0.85, 0.89, 0.95}

\definecolor{mediumtaupe}{rgb}{0.4, 0.3, 0.28}

\definecolor{bluegray}{rgb}{0.4, 0.6, 0.8}
\definecolor{celestialblue}{rgb}{0.29, 0.59, 0.82}
\definecolor{darkorange}{rgb}{1.0, 0.55, 0.0}

\definecolor{cadmiumred}{rgb}{0.89, 0.0, 0.13}
\definecolor{magnolia}{rgb}{0.97, 0.96, 1.0}
\definecolor{pastelblue}{rgb}{0.68, 0.78, 0.81}
\definecolor{persiangreen}{rgb}{0.0, 0.65, 0.58}
\definecolor{steelblue}{rgb}{0.27, 0.51, 0.71}

\definecolor{bluebell}{rgb}{0.64, 0.64, 0.82}
\definecolor{dimgray}{rgb}{0.41, 0.41, 0.41}
\definecolor{splashedwhite}{rgb}{1.0, 0.99, 1.0}

\definecolor{lavendergray}{rgb}{0.77, 0.76, 0.82}
\definecolor{lightgray}{rgb}{0.83, 0.83, 0.83}
\definecolor{lavendermist}{rgb}{0.9, 0.9, 0.98}

\definecolor{lightgreen}{HTML}{f8fcf4}
\definecolor{lightblue}{HTML}{dfebf7}
\definecolor{green_ncs}{rgb}{0.0, 0.62, 0.42}
\definecolor{islamicgreen}{rgb}{0.0, 0.56, 0.0}

\newcommand{\full}{\textsc{Visual Modality Instruction}}
\newcommand{\short}{\textsc{Vim}}

\newcommand{\tem}{\textsc{Tem}}

\newcommand{\vimmodel}{\textsc{v-MLLM}}

\newcommand{\llms}{LLMs}
\newcommand{\llm}{LLM}
\newcommand{\llmsfull}{Large Language Models}

\newcommand{\mllms}{MLLMs}
\newcommand{\mllmsfull}{Multimodal Large Language Models}

\title{Text as Images: Can Multimodal Large Language Models Follow Printed Instructions in Pixels?}

\author{Xiujun Li\textsuperscript{$^\spadesuit{}^*$}  
\quad Yujie Lu\textsuperscript{$^\clubsuit{}^*$} 
\quad Zhe Gan\textsuperscript{$^\diamondsuit$}
\quad Jianfeng Gao\textsuperscript{$^\heartsuit$} \\
\textbf{\quad William Yang Wang\textsuperscript{$^\clubsuit$}
\quad Yejin Choi\textsuperscript{$^\spadesuit$}} \\
\textsuperscript{$^\spadesuit$}University of Washington
\quad  
\textsuperscript{$^\clubsuit$}University of California, Santa Barbara \\
\textsuperscript{$^\diamondsuit$}Apple \quad \textsuperscript{$^\heartsuit$}Microsoft Research \\
\small{
\textsuperscript{$^*$}Equal contribution 
}
}

\begin{document}

\maketitle
\doparttoc 
\faketableofcontents 

\begin{abstract}
Recent multimodal large language models (\mllms) have shown promising instruction following capabilities on vision-language tasks. In this work, we introduce \full{} (\short)\footnote{\short{} is short for \textbf{\textsc{Vi}}sual \textbf{\textsc{m}}odality instruction, code: \url{https://github.com/VIM-Bench/VIM_TOOL}}, and investigate how well multimodal models can understand textual instructions provided in pixels, despite not being explicitly trained on such data during pretraining or fine-tuning. We adapt \short{} to eight benchmarks, including OKVQA, MM-Vet, MathVista, MMMU, and probe diverse \mllms{} in both the text-modality instruction (\tem) setting and \short{} setting. Notably, we observe a significant performance disparity between the original \tem{} and \short{} settings for open-source \mllms{}, indicating that open-source \mllms{} face greater challenges when text instruction is presented solely in image form. To address this issue, we train \vimmodel\footnote{\vimmodel{} is short for VIM-MLLM, model: \url{https://huggingface.co/VIM-Bench}}, a generalizable model that is capable to conduct robust instruction following in both text-modality and visual-modality instructions.
\end{abstract}

\section{Introduction}
\label{sec:intro}

Interleaved image-text data has been increasingly prevalent, ranging from web pages with images and tables, to user interfaces with instructions and forms, in which different modalities interact and blend together. For instance, to perform online shopping, an agent needs to understand the images, instructions and forms. Comprehensive understanding of this multi-modal data demands a range of skills, 
including recognizing text, understanding images, and also figuring out their interactions.

\begin{figure*}[!ht]
\centering
\includegraphics[width=\textwidth]{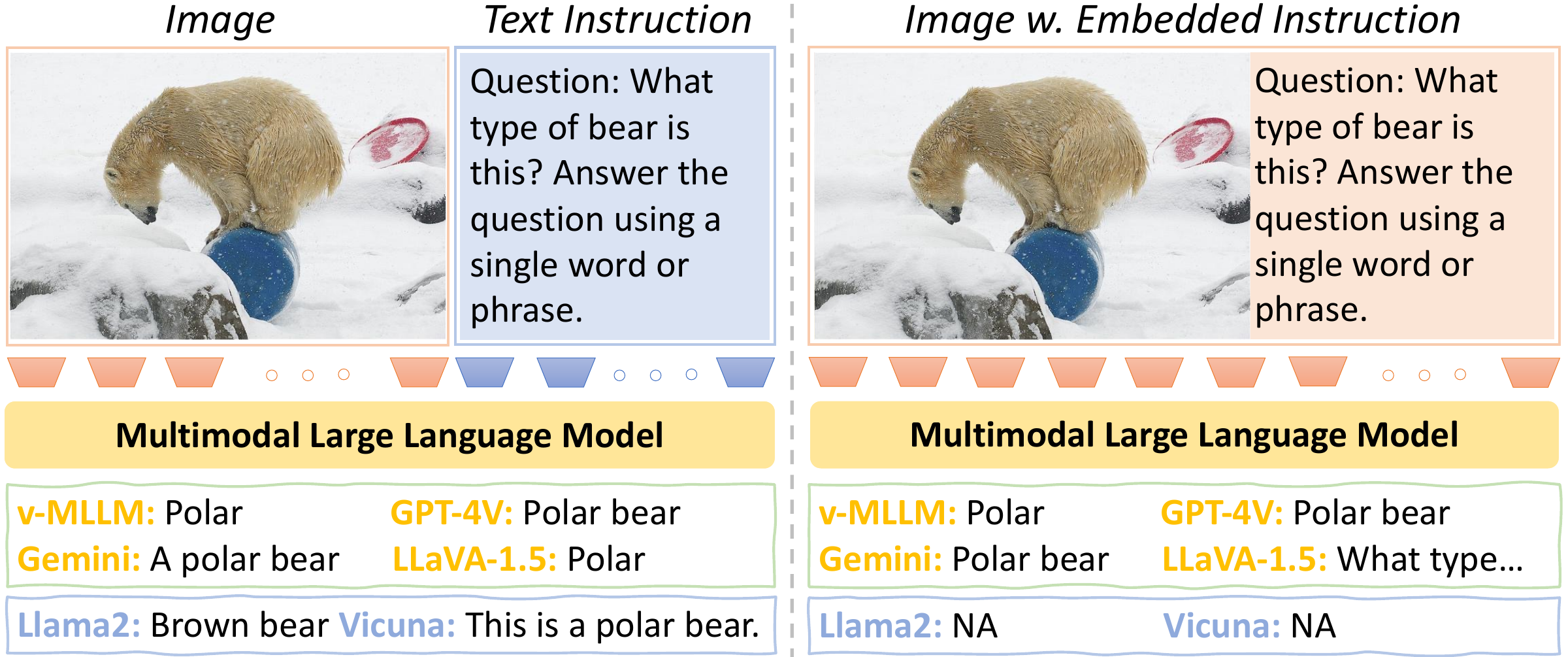}
\caption{
Evaluation paradigm comparison for \mllms. (a) Left is \tem{} setting, where \colorbox{macaroniandcheese}{Image} + \colorbox{wildblueyonder}{Text instruction} as two separate modalities are fed into \mllms{} for inference; an \llm{} model (for example, Vicuna) can also make correct prediction, even without accessing to the image. (b) Right: \short{} {\bf only} takes the \colorbox{macaroniandcheese}{image modality with the text instruction embedded in the image}, no additional text prompt is required, \llms{} are not applicable. The above example is from OKVQA (question \#209725). Note: \colorbox{macaroniandcheese}{Image modality input}, \colorbox{wildblueyonder}{Text modality input}.
}
\label{fig:vise_overview}
\end{figure*}

Inspired by the success of \llmsfull{} (\llms), the recent research on \mllmsfull{} (\mllms)~\cite{liu2023visual,dai2023instructblip,awadalla2023openflamingo,ye2023mplug,zhang2023llama,su2023pandagpt} may pave a way to understand this kind of vision-language data, and show promising results on a number of newly proposed benchmarks~\cite{fu2023mme,yu2023mm,liu2023mmbench,li2023seed,bitton2023visit,wu2023q}, demonstrating superb visual understanding, reasoning and generation capabilities. Despite their impressive performance,
they still remain poorly understood, and also these \mllms{} are not as impressive as their \llm{} counterparts, let alone their landing business applications. The current \mllms{} are built on top of the pretrained \llms, and \textit{visual} instruction tuning follows the recipe from its \llm{} counterparts, specifically, the instruction data is synthesized by the \llms{} (mostly from GPT-4 or GPT-4V) in the text format. As illustrated in the left part of Figure~\ref{fig:vise_overview}, the instruction and image are expressed in two modalities, we denote this kind of \textit{visual} instruction data as Text-Modality Instruction (\tem). Under this setting, a pure \llm, for example, Llama 2 or Vicuna in Figure~\ref{fig:vise_overview} can still make a plausible or correct prediction, even without accessing the image input. All the current benchmarks~\cite{fu2023mme,yu2023mm,liu2023mmbench,li2023seed,bitton2023visit} follow the same format.

\begin{wrapfigure}{r}{0.5\textwidth}
\centering
\includegraphics[width=0.5\textwidth]{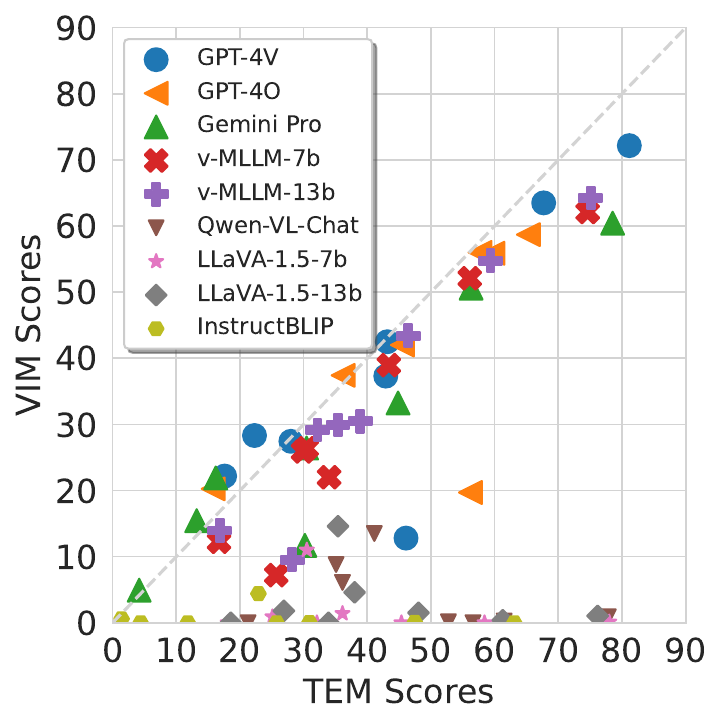}
\caption{
Model performance on the original \tem{} vs. \short{} settings. Each data point corresponds to one model's performance on one benchmark, in total, there are 72 data points (9 models * 8 tasks). This plot reveals: 1). Open-source \mllms{} (Qwen-VL-Chat, LLaVA, InstructBLIP) experience a significant performance drop from original \tem{} setting to \short{} setting; 2). GPT-4V, GPT-4O, Gemini Pro and our \vimmodel{} exhibit robust instruction following capability, as their data points are consistently align closely to the diagonal line.
} 
\label{fig:vise_radia_perf}
\end{wrapfigure}

This raises a question - \textit{how proficiently these \mllms{} can follow instructions if we embed the text instruction into visual format?} As shown in the right part of Figure~\ref{fig:vise_overview}, we name it as \full, where the image and instruction are in the visual modality. Under the \short{} setting, \llms{} are not applicable, and LLaVA-1.5 simply repeats the question for the image, may not understand the visual-modality instruction.

\begin{figure*}[t]
\centering
\includegraphics[width=\textwidth]{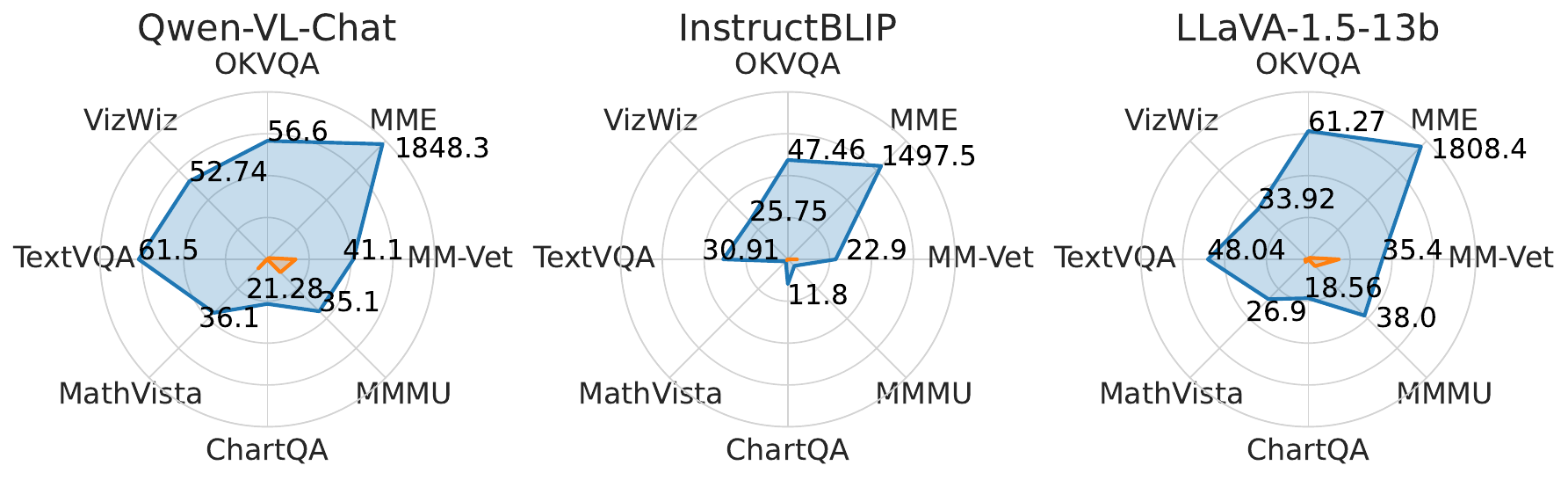}
\includegraphics[width=\textwidth]{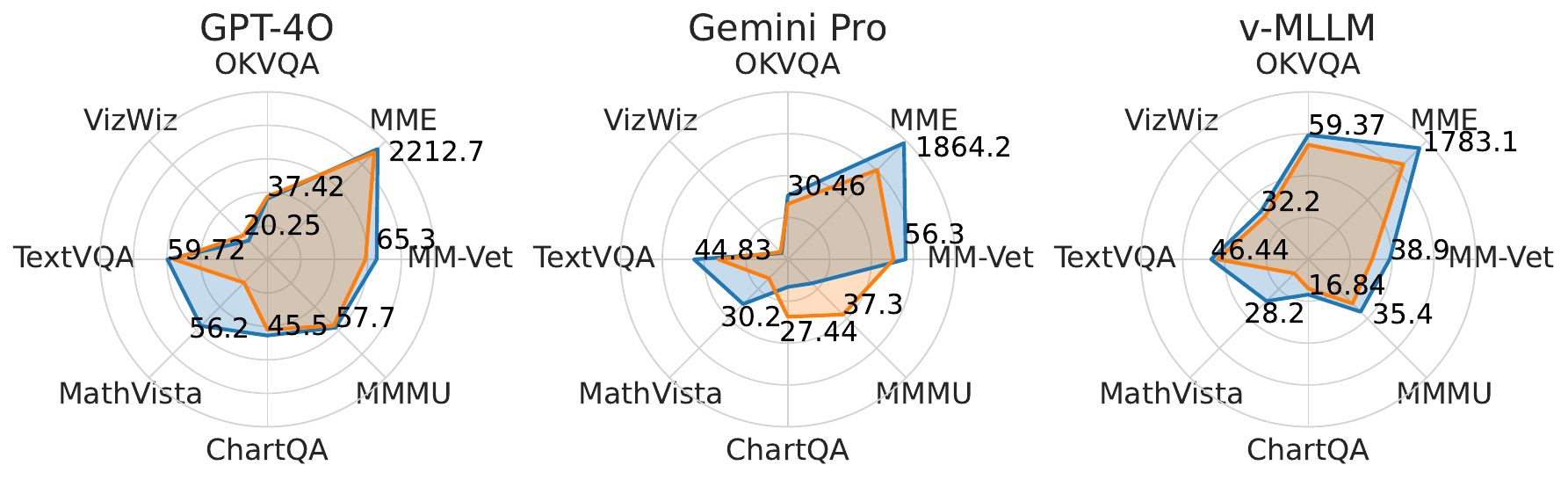}
\vspace{-5mm}
\caption{Performance comparison of six selected representative \mllms{} for visual instruction following between text-modality instruction (\textcolor{celestialblue}{TEM {\bf ---}}) and our introduced \full{} (\textcolor{darkorange}{\short{} {\bf ---}}) settings on eight benchmarks. There exists a disparity between \tem{} and \short{} settings for all open-source \mllms{} (the first row); GPT-4O, Gemini Pro and our \vimmodel{} are robust to instruction modality.}
\vspace{-2mm}
\label{fig:compare_radars}
\end{figure*}

Motivated by this, we introduce a new setting, called \full{} (short for \short), evaluating the capability of \mllms{} for visual-modality instruction following. We adapt \short{} to various benchmarks~\cite{marino2019ok,fu2023mme,yu2023mm,lu2023mathvista,yue2023mmmu}, and compose a new benchmark - \short{}-Bench. As highlighted in Figure~\ref{fig:vise_radia_perf} and \ref{fig:compare_radars}, there exists a performance disparity between the \tem{} and \short{} settings for all open-source \mllms{},
all of them are not 
robust enough at visual-modality instruction following.
To summarize, our main contributions are:
\begin{itemize}[itemsep=2pt,topsep=2pt,leftmargin=*] 
    \item We present \full{}, a challenging setting to probe the capability of \mllmsfull{} for visual-modality instruction following.
    \item We adapt the \short{} to various benchmarks, and reveal a significant disparity for open-source MLLMs between their text-modality instruction setting and \short{} setting.
    \item We train a \vimmodel{}, which demonstrates robust visual instruction following capabilities.
\end{itemize}
\section{Method}
\label{sec:vim}

Instruction following, is viewed as one key capability of high-performing \mllms{}. In this section, we first present \short{}, to examine the instruction following capability of \mllms{}, specifically the visual-modality instruction following. Then, we introduce \vimmodel, enhancing the \mllms{} with visual-modality instruction following.

\subsection{\short}

\subsubsection{Visual-Modality Instruction}
As illustrated in the left part of Figure~\ref{fig:vise_overview}, the current evaluation norm of \mllms{} takes two modalities as input: image and text (as instruction). The existing \mllms{} are built on top of the \llms{}, benefiting from its strong text understanding capability. For the current MLLM evaluation paradigm, instruction is presented in the text modality, which can utilize the strong language priors from the \llms{} for understanding. As shown in Table~\ref{tab:two_results}, even a pure \llm{} model (GPT-4, Llama 2 or Vicuna) can get some success without accessing to the images. Interestingly, on most of eight tasks, Llama 2 shows better numbers over GPT-4 (\texttt{gpt-4-1106-preview}). We manually check some response, and find that the output from GPT-4 are more reasonable than Llama 2. This might rise several interesting issues, we leave a discussion in Section~\ref{sec:llm_response}.

\short{} challenges the \mllms{} by rendering the textual instruction into the visual pixel space (image), this enhancement demands not just textual but also strong visual comprehension for instruction understanding. It asks for the strong visual interpretation capability to recognize and follow the embedded instruction in the image.

\subsubsection{Design Choices}
How to lay out the image and embedded instruction in one visual space? For zero-shot setting, there are many combinations to position the instruction and image in the same visual space. Here, we enumerate two important elements we investigated.

\begin{wrapfigure}{r}{0.5\textwidth}
\centering
\includegraphics[width=0.49\textwidth]{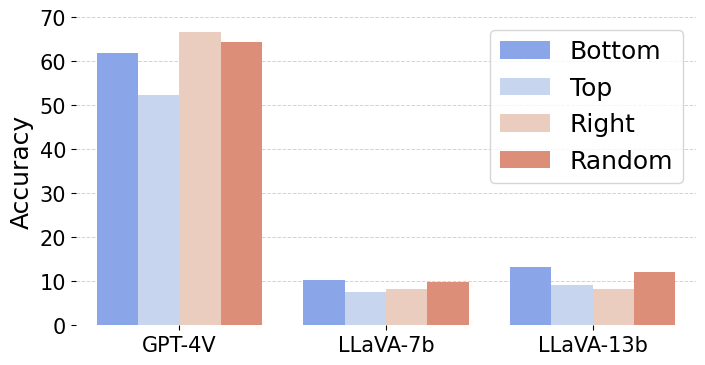}
\caption{Exploration setup for instruction location on zero shot evaluation for MM-Vet.}
\label{fig:loc_exp}
\vspace{-3mm}
\end{wrapfigure}

\paragraph{Instruction Location} Potentially, there are many options to place the instruction into the image. To narrow down the search choices of instruction's placement, we focus on three primary positions: \{top, right and bottom\}. Additionally, we add a random selection from these three positions to ensure robustness. Preliminary experiments\footnote{Here the small subset for preliminary experiments is $21$ examples from MM-Vet.} on these locations reveal that both GPT-4V and LLaVA-1.5 are robust to the locations of the embedded instruction, as shown in Figure~\ref{fig:loc_exp}. For the sake of simplicity, we take the bottom position as the default placement for the embedded instruction.

\paragraph{Image Resolution}
The resolution of the image is also the key for the model to understand the visual-modality instruction. For the image-text pairs, we aim to keep the resolution of the raw images, we add the text with the same font size for all images and add extra white space fillings to render the text regarding the length of the instruction. The resolution of image is minimally changed, and the origin image quality is maintained.

\subsubsection{Text Prompting}
To probe the visual instruction following of \mllms{}, we prompt the models under three settings: 1) \texttt{Text-Modality Instruction} (\tem): given the image, the instruction is fed to the models as prompt in text format; 2) \texttt{Mix Instruction}: given the image with embedded instruction, an extra instruction as text prompt can be taken as input to guide models to follow the embedded instruction in the image, e.g., ``\textit{Answer the question in the image}''; 3) \texttt{Visual-Modality Instruction}: given the image with the embedded instruction, no extra text instruction is provided as text prompt, the model needs to recognize, and understand the embedded instruction in the image automatically and follow the instruction to deliver the answer.

\begin{wraptable}{r}{0.5\textwidth}
\caption{Exploration setup for text prompt on zero-shot evaluation for MM-Vet. * denotes from the paper reported~\cite{liu2023improved}. }
\label{tab:text_prompt_exp}
\resizebox{0.5\textwidth}{!}{
\begin{tabular}{l@{\hspace{5pt}}c|@{\hspace{5pt}}ccc}
\toprule
Models & LLM & \texttt{\tem} & \texttt{Mix} & \texttt{\short}  \\ 
\midrule
\multicolumn{5}{c}{\textit{Small set}} \\
\midrule
GPT-4V & - & $66.7$ & $54.4$  & $63.5$  \\
\midrule
\multicolumn{5}{c}{\textit{Full Set}} \\
\midrule
LLaVA-1.5 & {\small Vicuna-7b} & $30.5^*$ & $10.3$ ($\textcolor{cadmiumred}{\bf -20.2}$) & $11.0$ ($\textcolor{cadmiumred}{\bf -19.5}$) \\
LLaVA-1.5 & {\small Vicuna-13b} & $35.4^*$ & $14.8$ ($\textcolor{cadmiumred}{\bf -20.6}$) & $14.6$ ($\textcolor{cadmiumred}{\bf -20.8}$) \\
InstructBLIP & {\small FlanT5$_\text{XXL}$} & $22.9$ & $12.5$ ($\textcolor{cadmiumred}{\bf -10.4}$) & $4.4$ ($\textcolor{cadmiumred}{\bf -18.5}$) \\
Qwen-VL-Chat & - & $41.1$ & $26.2$ ($\textcolor{cadmiumred}{\bf -14.9}$) & $13.5$ ($\textcolor{cadmiumred}{\bf -27.6}$) \\
\bottomrule
\end{tabular}
}
\end{wraptable} 

The \texttt{Text-Modality Instruction} setting is the standard setup in \mllms{}, that the model will take the image and text question or instruction separately through vision and language encoders respectively. In the \texttt{Mix Instruction} and \texttt{Visual-Modality Instruction} settings, the models are required to understand the embedded instruction in the image, while, \texttt{Mix Instruction} is a relaxed setting between \texttt{Text-Modality Instruction} and \texttt{Visual-Modality Instruction}. Preliminary results in Table~\ref{tab:text_prompt_exp} demonstrates that GPT-4V is robust to all three prompt settings, while, \short{} is more challenging for the existing open-source \mllms{}, for example, LLaVA-1.5 and Qwen-VL-Chat drop more ($>$$20$) when transferring from \texttt{Text-Modality Instruction} to the \short{} setting, similarly for InstructBLIP. In this paper, we will use the \short{} setting for the rest of experiments.

\subsection{\vimmodel}
\label{sec:vim_mllm}

\subsubsection{\short{} Corpus}
One key ingredient of high-performing \mllms{} is high-quality instruction tuning data. There are two categories of visual instruction tuning data, one is the synthetic data by \llms{} (i.e. GPT-4), like LLaVA~\cite{liu2023visual}; the other one is the synthetic data generated by GPT-4V, like LVIS-Instruct4V~\cite{wang2023see} and ShareGPT-4V~\cite{chen2023sharegpt4v}. Here, we use the LVIS-Instruct4V-LLaVA-Instruct-mix880k~\cite{wang2023see}
as our origin instruction tuning corpus $D_R$, and convert it into the \short{} format. 
We only consider the first turn for the multiple turn conversation data. In total, we get 846k \short{} training data $D_V$ after filtering the unavailable image links.

\subsubsection{\short{} Training}
\vimmodel{} adopts a similar architecture with LLaVA-1.5~\cite{liu2023improved} and LVIS-Instruct4V~\cite{wang2023see}. The model follows an autoregressive training approach, focusing on optimizing the sequential prediction of the answer words $y_1, y_2, \ldots, y_n$ by minimizing the loss function 
$$
L = \sum_{i=1}^{n} \text{loss}(\text{LM}(y_{<i}, T, V), y_i)
$$
where $y_{<i}$ signifies all tokens preceding the $i$-th token, $T$ and $V$ represent the textual (e.g., text-modality instruction or prompt) and visual (e.g., visual-modality instruction and image context) tokens. Here the textual token sequence $T$ is optional in the \short{} training.


To train a unified model that can robustly follow the text-modality and visual-modality instructions, there are two strategies. 
\begin{itemize} [noitemsep,topsep=2pt,leftmargin=*] 
    \item \texttt{Stage-wise} training, is to train the \vimmodel{} on the origin corpus $D_R$ first, then continue to train the model on the \short{} corpus $D_V$.
    \item \texttt{Mixture} training, is to combine the original corpus $D_R$ and \short{} corpus $D_V$ as one corpus $D = \{D_R, R_V\}$, and train the model with in-batch random sampling. 
\end{itemize}
For the \texttt{stage-wise} training, there might be one potential issue - catastrophic forgetting, the model may forget some of its behaviors after the second stage of training. Empirically, we compared the two training strategies and found there is no significant difference in Section~\ref{sec:ab_training}. Following the architectures of vicuna~\cite{vicuna2023} and LLaVA~\cite{liu2023visual}, we train two versions of the model on the top of LVIS-Instruct4V~\cite{wang2023see}: \vimmodel-7B and \vimmodel-13B.

\begin{table}[t]
\caption{Main quantitative results over each benchmark under \textit{\tem} and \textit{\short{}} settings. \colorbox{isabelline}{\makebox(5,5){}}: \llm{} models, \colorbox{magnolia}{\makebox(5,5){}}: proprietary models, \colorbox{lightblue}{\makebox(5,5){}}: the proposed models. *We use a more strict evaluation protocol to remove randomness when mapping from open-ended responses to multiple choices.
}
\label{tab:two_results}
\vspace{-3mm}
\begin{center}
\resizebox{\linewidth}{!}{
\begin{tabular}{l@{\hspace{3pt}}@{\hspace{3pt}}c@{\hspace{6pt}}c|c@{\hspace{0pt}}c@{\hspace{5pt}}c@{\hspace{5pt}}c@{\hspace{5pt}}c@{\hspace{5pt}}c@{\hspace{5pt}}c@{\hspace{5pt}}c}
\toprule
Models & LLM & Res. & MM-Vet & MME & OKVQA & VizWiz & TextVQA & MathVista${^*}$ & ChartQA & MMMU${^*}$  \\ \midrule
\multicolumn{11}{c}{\textit{\tem{} Setting}} \\
\rowcolor{isabelline} GPT-4 & - & - & $9.8$ & $74.6$ & $8.37$ & $2.76$ & $3.36$ & $18.7$ & $4.12$ & $28.8$ \\
\rowcolor{isabelline} Llama 2 & {\footnotesize Llama2-7b} & - & $11.1$ & $1609.5$ & $16.21$ & $5.67$ & $7.18$ & $23.2$ & $0$ & $6.2$ \\
\rowcolor{isabelline} Vicuna & {\footnotesize Vicuna-7b} & - & $11.7$ & $1120.6$ & $4.5$ & $1.88$ & $1.88$ & $18.1$ & $0$ & $2.0$
\vspace{2mm}
\\
\rowcolor{magnolia} GPT-4V & - & - & $\bf 67.7$ & $1926.6$ & $22.28$ & $17.59$ & $43.14$ & $46.1$ & $28.00$ & $42.9$ \\
\rowcolor{magnolia} GPT-4O & - & - & $65.3$ & $\bf 2212.7$ & $36.20$ & $15.79$ & $59.72$ & $\bf 56.2$ & $\bf 45.50$ & $\bf 57.7$ \\
\rowcolor{magnolia} Gemini Pro & - & - & $56.3$ & $1864.2$ & $30.46$ & $4.17$ & $44.83$ & $30.2$ & $13.20$ & $16.2$\\
Qwen-VL-Chat & - & - & $41.1$ & $1848.3$ & $56.6$ & $\bf 52.74$ & $\bf 61.5$ & $36.1$ & $21.28$ & $35.1$ \\
InstructBLIP & {\footnotesize FlanT5$_\text{XXL}$} & $224$ & $22.9$ & $1497.5$ & $47.46$ & $25.75$ & $30.91$ & $1.4$ & $11.80$ & $4.40$ \\
LLaVA-1.5 & {\footnotesize Vicuna-7b} & $336$ & $30.5$ & $1851.5$ & $58.41$ & $32.08$ & $45.36$ & $25.1$ & $18.08$ & $36.1$ \\
LLaVA-1.5 & {\footnotesize Vicuna-13b} & $336$ & $35.4$ & $1808.4$ & $\bf 61.27$ & $33.92$ & $48.04$ & $26.9$ & $18.56$ & $38.0$ \\
\rowcolor{lightblue} 
\vimmodel & {\footnotesize Vicuna-7b} & $336$ & $29.9$ & $1771.1$ & $56.09$ & $30.48$ & $43.38$ & $25.7$ & $16.72$ & $34.0$ \\
\rowcolor{lightblue} 
\vimmodel & {\footnotesize Vicuna-13b} & $336$ & $38.9$ & $1783.1$ & $59.37$ & $32.20$ & $46.44$ & $28.2$ & $16.84$ & $35.4$ \\
\midrule
\multicolumn{11}{c}{\textit{\short{} Setting}} \\
\rowcolor{magnolia} GPT-4V & - & - & $\bf 63.5$ & $1713.1$ & $28.32$ & $22.18$ & $42.50$ & $12.8$ & $27.44$ & $37.3$ \\
\rowcolor{magnolia} GPT-4O & - & - & $58.7$ & $\bf 2144.3$ & $37.42$ & $20.25$ & $\bf 55.88$ & $\bf 19.7$ & $\bf 42.00$ & $\bf 56.0$ \\
\rowcolor{magnolia} Gemini Pro & - & - & $50.6$ & $1434.6$ & $26.43$ & $4.93$ & $33.24$ & $11.7$ & $15.44$ & $21.9$ \\
Qwen-VL-Chat & - & - & $13.5$ & $21.2$ & $0.01$ & $0.15$ & $0.27$ & $6.1$ & $0$ & $8.8$ \\
InstructBLIP & {\footnotesize FlanT5$_\text{XXL}$} & $224$ & $4.40$ & $0$ & $0.07$ & $0$ & $0.04$ & $0.6$ & $0$ & $0$ \\
LLaVA-1.5 & {\footnotesize Vicuna-7b} & $336$ & $11.0$ & $2.9$ & $0$ & $0$ & $0$ & $0.9$ & $0$ & $1.4$ \\
LLaVA-1.5 & {\footnotesize Vicuna-13b} & $336$ & $14.6$ & $24.4$ & $0.38$ & $0$ & $1.51$ & $1.8$ & $0$ & $4.6$ \\
\rowcolor{lightblue} 
\vimmodel & {\footnotesize Vicuna-7b} & $336$ & $25.9$ & $1474.6$ & $52.10$ & $26.40$ & $38.96$ & $7.2$ & $12.24$ & $22.0$ \\
\rowcolor{lightblue}
\vimmodel & {\footnotesize Vicuna-13b} & $336$ & $30.5$ & $1525.1$ & $\bf 54.76$ & $\bf 29.15$ & $43.40$ & $9.5$ & $13.96$ & $29.9$ \\
\bottomrule
\end{tabular}
}
\end{center}
\end{table}

\section{Experiments}
\label{sec:exp}
We first build our \short{}-Bench based on eight existing representative benchmarks, then compare the \vimmodel{} with six representative \mllms{} 
under two settings (\tem{} and \short) across all the tasks.

\subsection{\short{}-Bench}
\label{sec:vim_bench}
\paragraph{Benchmarks} To assess the generalization capability of \mllms, we adapt \short{} to eight representative benchmarks, including 
MME~\cite{fu2023mme}, MM-Vet~\cite{yu2023mm}, OKVQA~\cite{marino2019ok}, VizWiz~\cite{bigham2010vizwiz}, TextVQA~\cite{singh2019towards}, MathVista~\cite{lu2023mathvista}, ChartQA~\cite{masry2022chartqa}, and MMMU~\cite{yue2023mmmu}. 
The details of source datasets, data processing pipeline, and evaluations can be found in Appendix~\ref{sec:benchmark_details}.

\paragraph{Data Reformatting}
Given the above mentioned benchmarks, we try to do minimal changes (i.e., keeping the image resolution) for evaluation. This process involves reformatting text instruction into visual-modality instruction by moving the text instruction into the image modality. In reformatting, we retain the original task’s goal while maintaining the original images with text instructions rendering at the bottom of the image, see the example in Figure~\ref{fig:vise_overview}. These repurposed benchmarks are integrated into our \short{}-Bench. Theoretically, \short{} can be applied to any existing benchmarks, even for pure NLP tasks. We choose eight representative MLLM benchmarks, although our selections are not exhaustive, they provide a broad basis for MLLM evaluation. 

\subsection{Evaluation Setup}

\paragraph{Evaluation Settings} For each benchmark, we have two main evaluation settings, one is the standard evaluation setting, which is Text-Modality Instruction, \texttt{image} + \texttt{text prompt} as the input, denoted as \textit{\tem} setting; the other one is the \textit{\short} setting with only \texttt{image} as input. For the \short{} setting, it also can accept any text prompts, we call \texttt{Mix Instruction} in the ablation experiments.

\paragraph{Evaluation Metrics}
For the evaluation metrics, we follow the evaluation protocols and metrics of each benchmark. For MathVista and MMMU, the open-ended responses will be parsed into options before calculating the accuracy. We take a more strict protocol that requires the models to follow explicit instructions (e.g., ``Answer with the choice letter.''), instead of mapping to the closest options no matter whether the response is related or not. Details of metrics are illustrated in Appendix~\ref{sec:app_evaluation_detail}.

\subsection{Baselines}
\label{sec:baselines}
We evaluate nine representative open-source and proprietary \mllms{} (with different variations), including InstructBLIP~\cite{dai2023instructblip}, LLaVA-1.5~\cite{liu2023improved}, Qwen-VL-Chat~\cite{bai2023qwenvl}, 
GPT-4V~\cite{gpt4}, Gemini Pro~\cite{team2023gemini}, and latest GPT-4O~\cite{gpt-4o}. For LLaVA-1.5, we use its latest versions with Vicuna-v1.5 as the LLM backbone. We use Flan-T5 XXL~\cite{chung2022scaling} as LLM backbone for InstructBLIP. 
All experiments are conducted on NVIDIA A100 and H100 80G GPUs, we use as large models as possible, model training was integrated with Deepspeed Zero-3. For 7b size models, we use batch size 80 with 3 epochs; for 13b models, batch size is 48 with 2 epochs on a single node with 8 GPUs.

\subsection{Main Results}
Table~\ref{tab:two_results} summarizes the overall results for two settings. 1). In the original \tem{} setting, the backbone \llm{} models can get decent performance on these benchmarks, even without access to the image modality. Interestingly, on six out of eight tasks, Llama 2 is much better than GPT-4, we will briefly discuss this issue in Section~\ref{sec:llm_response}. 2). For all open-source \mllms, there is a significant performance disparity between the \tem{} setting and \short{} setting. 3). GPT-4V and Gemini Pro are robust to the instruction modality, while, open-source \mllms{} struggle in the \short{} setting, achieve significantly low scores. 4). Our proposed \vimmodel{} shows robust instruction following capabilities in two settings across all the tasks, especially in the \short{} setting, significant gain over open-source \mllms.

\begin{table}[t!]
\caption{MLLMs' instruction recognition response to the question \#3575865 in OKVQA.}
\label{tab:ab_fs_example}
\vspace{-3mm}
\centering
\begin{tcolorbox}[colback=splashedwhite,boxsep=1pt,left=2pt,right=5pt,top=5pt,bottom=3pt,boxrule=0.5pt]
\centering
\small
\begin{tabular}{p{0.38\columnwidth} p{0.55\columnwidth}}
\VarSty{ {\bf {\textit{Image w. Embedded Instruction}}} } & \VarSty{{\bf {\textit{Text Prompt}}}}: What is the text in the image? \\ 
\\
\vspace{-6mm}
\multirow{4}{*}{
\includegraphics[height=3.9cm]{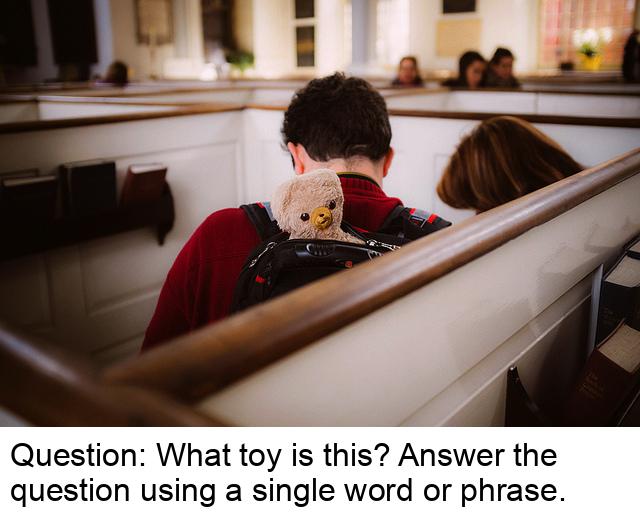} 
}  &  \multicolumn{1}{c}{\bf \texttt{Recognized Instructions}} \\
\\
& \VarSty{ {\bf LLaVA-1.5-13B:}} The image shows a man sitting in a pew with a teddy bear on his back. The teddy bear is wearing a backpack, and the man appears to be looking at it. The scene takes place in a church, with several other people present in the background. \\ 
\\
& \VarSty{ {\bf GPT-4V: }} The text in the image says:
"Question: What toy is this? Answer the question using a single word or phrase."  \\
\end{tabular}
\end{tcolorbox}
\end{table}

\section{Ablation}
\label{sec:ablation}

We have demonstrated that \short{} is a challenging setting for the current open-source \mllms{}. To better understand the gap between the standard \texttt{Text-Modality Instruction} and \texttt{Visual-Modality Instruction}, we decompose the \short{} into two steps for ablation.

\begin{wrapfigure}{r}{0.6\textwidth}
\centering
\includegraphics[width=0.6\textwidth]{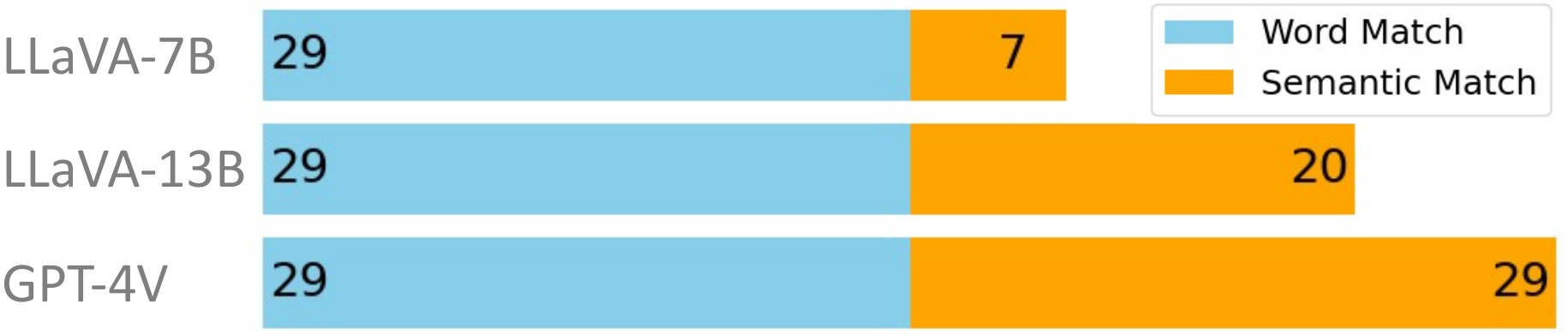}
\caption{Instruction recognition results on VQA$_{v2}$. We report the correct number of matches out of total 30 samples for \textcolor{gray}{LLaVA-1.5-7B \& 13B}, and \textcolor{gray}{GPT-4V}.
}
\label{fig:ocr_results}
\end{wrapfigure}

\subsection{Instruction Recognition}
\label{sec:instr_rec}
One hypothesis is the vision encoder of \mllms{} cannot discern the instruction in the image, only generates the long description for the image, as the LLaVA example shown in Table~\ref{tab:ab_fs_example}. \textit{Can the \mllms{} recognize the embedded instruction in the image?} Here, we conduct an experiment to verify the visual-modality instruction recognition ability of the models, we choose an external dataset VQA$_{v2}$ for test (beyond our eight test benchmarks)\footnote{Due to the large size of VQA$_{v2}$, it is expensive to run the full \textit{test-dev} and \textit{test} splits for GPT-4V and Gemini Pro for fair comparison, we did not include it in the \short{}-Bench.}, we explicitly prompt the models with ``\textit{What is the text in the image?}'', then manually check the outputs with the ground-truth text instructions in the zero-shot setting for VQA$_{v2}$. We select $30$ images as our subsets for both models. Here, the images we choose only have the embedded instructions since we want to verify the instruction recognition capability of these models in an ideal setup.

We do \texttt{word match} and \texttt{semantic match} for the results. For example, the origin instruction of question \#$393225000$ in VQA$_{v2}$ is ``\textit{What website copyrighted the picture}'', the output ``\textit{What website copied the picture?}'' is counted as a correct word match, since most of the words are recognized. However, it will be considered as a wrong semantic match. As shown in Figure~\ref{fig:ocr_results}, GPT-4V can recognize the embedded instructions nearly perfectly on word match and semantic match, the only ``\textit{failure}'' case is from an image containing a text logo inside as shown in Table~\ref{tab:ab_instr_rec} (in the Appendix). While, LLaVA-1.5 can detect words of the instructions, but semantically different. A detailed analysis of instruction recognition capability can be found in Appendix~\ref{sec:app_inst_rec}.

\begin{table}[t]
\caption{Ablation results on OKVQA, MM-Vet, VizWiz tasks. 
}
\label{tab:ab_mix_result}
\vspace{-2mm}
\begin{center}
\resizebox{0.9\textwidth}{!}{
\begin{tabular}{c@{\hspace{5pt}}c@{\hspace{5pt}}|cc|cc|cc}
\toprule
\multirow{2}{*}{Models} & \multirow{2}{*}{LLM} & \multicolumn{2}{c|}{OKVQA} & \multicolumn{2}{c|}{MM-Vet} & \multicolumn{2}{c}{VizWiz} \\
 &  & \texttt{VIM} & \texttt{MIX} & \texttt{VIM} & \texttt{MIX} & \texttt{VIM} & \texttt{MIX} \\
\midrule
LLaVA-1.5 & Vicuna-7b  & $0.00$ & $14.28$ (\textcolor{islamicgreen}{+$14.28$}) & $11.0$ & $10.3$ (\textcolor{red}{-$0.7$}) & $0$ & $22.47$ (\textcolor{islamicgreen}{+$22.47$}) \\
InstructBLIP &    FlanT5$_\text{XXL}$     & $0.07$ & $25.44$ (\textcolor{islamicgreen}{+$25.37$})& $4.4$ & $12.5$ (\textcolor{islamicgreen}{+$8.1$}) & $0$ & $18.79$ (\textcolor{islamicgreen}{+$18.79$}) \\
Qwen-VL-Chat &    -    & $0.01$ & $30.75$ (\textcolor{islamicgreen}{+$30.74$}) & $13.5$ & $26.2$ (\textcolor{islamicgreen}{+$12.7$}) & $0.15$ & $34.03$ (\textcolor{islamicgreen}{+$33.88$}) \\

GPT-4V & - & $28.32$ & $27.70$ (\textcolor{red}{-$0.62$}) & $63.5$ & $54.4$ (\textcolor{red}{-$9.1$}) & $22.18$ & $23.37$ (\textcolor{islamicgreen}{+$1.19$}) \\
\vimmodel & Vicuna-7b & $52.10$  & $51.82$ (\textcolor{red}{-$0.28$}) & $25.9$ & $24.4$ (\textcolor{red}{-$1.5$}) & $26.40$ & $27.34$ (\textcolor{islamicgreen}{+$0.94$}) \\
\bottomrule
\end{tabular}
}
\end{center}
\end{table}

\subsection{Instruction Following}
\label{sec:ab_instr_follow}
We further relax the \short{} setting to the \texttt{Mix Instruction} setting with an extra text prompt ``\textit{Answer the question in the image in one word or phrase.}'', these open-source \mllms{} exhibit some degree of success (the highlighted \textcolor{islamicgreen}{green} columns in Table~\ref{tab:ab_mix_result}), 
which proves that the existing \mllms{} rely more on their LLM backbones for instruction following. For example, on OKVQA and VizWiz, LLaVA-1.5, InstructBLIP and Qwen-VL-Chat can achieve moderate success (\textcolor{islamicgreen}{+10}) in the \texttt{Mix Instruction} setting, significant improvement over their \short{} setting.

\subsection{Mixture training v.s. Stage-wise training}
\label{sec:ab_training}
We ablation the two training strategies on the whole corpus, and verify the results on three downstream tasks under both the \tem{} and \short{} settings. Table~\ref{tab:ab_training_res} shows that 1). Training procedure is often unstable, hard to seek a well balanced checkpoint on all the tasks, which is consistent with the observation in screenshot LM~\cite{gao2024improving}. 2). There is no significant difference for mixture training and stage-wise training, we use stage-wise training in our experiments. 3). Referring to LLaVA, \vimmodel{} maintains comparable performance under the \tem{} setting and achieves significantly better performance under the \short{} setting.

\begin{table}[t]
\caption{Ablation results on training strategies. Note: Taking LLaVA as referring baseline. 
}
\vspace{-3mm}
\label{tab:ab_training_res}
\begin{center}
\resizebox{\textwidth}{!}{
\begin{tabular}{c@{\hspace{5pt}}c@{\hspace{5pt}}c|cc|cc|cc}
\toprule
\multirow{2}{*}{Models} & \multirow{2}{*}{LLM} & \multirow{2}{*}{Training} & \multicolumn{2}{c|}{OKVQA} & \multicolumn{2}{c|}{MM-Vet} & \multicolumn{2}{c}{TextVQA} \\
 &  &  & \texttt{TEM} & \texttt{VIM} & \texttt{TEM} & \texttt{VIM} & \texttt{TEM} & \texttt{VIM} \\
\midrule

\textcolor{gray}{LLaVA-1.5} & \textcolor{gray}{Vicuna-7b} & \textcolor{gray}{-} & \textcolor{gray}{$58.41$} & \textcolor{gray}{$0$} & \textcolor{gray}{$30.5$} & \textcolor{gray}{$11.0$} & \textcolor{gray}{$45.36$} & \textcolor{gray}{$0$} \\

\vimmodel & Vicuna-7b & Stage-wise & $56.09$ & $52.10$ & $29.9$ & $25.9$ & $43.38$ & $38.96$ \\
\vimmodel & Vicuna-7b & Mixture & $58.74$ (\textcolor{islamicgreen}{+$2.65$}) & $52.90$ (\textcolor{islamicgreen}{+$0.8$}) & $28.8$ (\textcolor{red}{-$1.1$}) & $23.5$ (\textcolor{red}{-$2.4$}) & $45.49$ (\textcolor{islamicgreen}{+$2.11$}) & $41.77$ (\textcolor{islamicgreen}{+$2.81$}) \\

\textcolor{gray}{LLaVA-1.5} & \textcolor{gray}{Vicuna-13b} & \textcolor{gray}{-} & \textcolor{gray}{$61.27$} & \textcolor{gray}{$0.38$} & \textcolor{gray}{$35.4$} & \textcolor{gray}{$14.6$} & \textcolor{gray}{$48.04$} & \textcolor{gray}{$1.51$} \\

\vimmodel & Vicuna-13b & Stage-wise & $59.37$ & $54.76$ & $38.9$ & $30.5$ & $46.44$ & $43.40$ \\
\vimmodel & Vicuna-13b & Mixture & $57.84$ (\textcolor{red}{-$1.53$}) & $53.09$ (\textcolor{red}{-$1.67$}) & $37.6$ (\textcolor{red}{-$1.2$}) & $28.6$ (\textcolor{red}{-$1.9$}) & $45.77$ (\textcolor{red}{-$0.67$}) & $43.42$ (\textcolor{islamicgreen}{+$0.02$}) \\
\bottomrule
\end{tabular}
}
\end{center}
\end{table}

\subsection{Qualitative Observations}
\label{sec:qual_ob}
We highlight several intriguing observations during the experiments, and a comprehensive qualitative analysis is provided in the Appendix~\ref{sec:app_qual_ob}.

\section{Discussion}
\label{sec:discussion}

\begin{table}[t!]
\caption{\llms' responses to the questions in MM-Vet and OKVQA. {\small (\VarSty{{\bf {\textit{Image}}}} is not feed into the \llms.)}}
\label{tab:llm_exp_output}
\vspace{-3mm}
\centering
\begin{tcolorbox}[colback=splashedwhite,boxsep=1pt,left=2pt,right=5pt,top=5pt,bottom=3pt,boxrule=0.5pt]
\centering
\small
\begin{tabular}{p{0.38\columnwidth} p{0.55\columnwidth}}
\multicolumn{2}{c}{\bf \texttt{MM-Vet \#1}} \\
\VarSty{ {\bf {\textit{Image}}}}: 
& \VarSty{ {\bf {\textit{Text Prompt}}}}: What is d in the last equation? Answer the question in one word or phrase. \\ 
\\
\vspace{-10mm}
\multirow{4}{*}{
\includegraphics[height=2.6cm]{ab_imgs/v1_1.png}
}  &  \VarSty{ {\bf GPT-4:}} I'm sorry, but I need more context or information about the "last equation" you are referring to in order to provide an answer. Could you please provide the equation or more details surrounding it?  \\
& \VarSty{ {\bf Vicuna:}} d is the derivative of the function with respect to x. \\ 
& \VarSty{ {\bf Llama 2:}} D is the constant of integration. \\
\\
\multicolumn{2}{c}{\bf \texttt{OKVQA \#2971475}} \\
\vspace{-3mm}
\VarSty{ {\bf {\textit{Image}}}}: 
& \VarSty{{\bf {\textit{Text Prompt}}}}: What sport can you use this for? Answer the question in one word or phrase. \\ 
\\
\vspace{-12mm}
\multirow{4}{*}{\qquad\quad
\includegraphics[height=2.5cm]{ab_imgs/COCO_val2014_000000297147.png}
}  &  \VarSty{ {\bf GPT-4:}} Please provide the context or the object you're referring to in order to answer the question about which sport it can be used for.   \\
& \VarSty{ {\bf Vicuna:}} Basketball. \\ 
& \VarSty{ {\bf Llama 2:}} Basketball 
\end{tabular}
\end{tcolorbox}
\end{table}

GPT-4V, GPT-4O and Gemini Pro demonstrate robust instruction following under the original \tem{} and \short{} settings, especially in the \short{} setting, they show robust response, while, most of \mllms{} struggle under the \short{} setting. Our proposed \vimmodel{} narrows down this gap for the open-source \mllms, and bring robust instruction following capability to different instruction modalities. There could be several directions to be explored along this routine. One direction is integrating \full{} with grounding abilities of \mllms{} discussed by several works~\cite{yang2023set,chen2023shikra,you2023ferret}. Another direction is scaling up, in terms of data scale and model scale.

\paragraph{MLLM Evaluation}
\label{sec:llm_response}
In the \llm{} exploration setting, Table~\ref{tab:two_results} shows that Llama 2 is much better than GPT-4 on six of eight multimodal tasks, without image input. We manually check some results, and find that the responses from GPT-4 are more reasonable than Vicuna and Llama 2. For example, in Table~\ref{tab:llm_exp_output}, we only provide the text \texttt{Prompt} to the \llms, the output from Vicuna and Llama 2 is more like text continuation based on the training corpus, while, GPT-4 makes more reasonable response to the \texttt{Prompt} question. 

Another potential issue exposed from this exploration experiment is about MLLM evaluation, though GPT-4 makes more reasonable response, its score is low on all six tasks, which may be contrary to the objective of these benchmarks. In-depth analysis and discussion of evaluation protocol and metrics are beyond the scope of this work, which may leave for future work.
\section{Related Work}
\label{sec:related}

\paragraph{\mllmsfull} With the success of Large Language Models (\llms)~\cite{gpt4,google2023bard,touvron2023llama,touvron2023llama2}, there is growing interest in studying \mllmsfull~\cite{liu2023visual,dai2023instructblip,zhu2023minigpt,awadalla2023openflamingo,ye2023mplug,zhang2023llama,su2023pandagpt,sun2023aligning,bai2023qwenvl} to improve multimodal understanding, reasoning and generation by leveraging the strong capability of \llms. Following the recipe of the instruction tuning~\cite{taori2023stanford,vicuna2023,peng2023instruction} in \llms, LLaVA~\cite{liu2023visual} and MiniGPT-4~\cite{zhu2023minigpt} propose to use GPT-4~\cite{gpt4} to synthesize the instruction, and employ the open-source \llms{} for instruction tuning to connect the pretrained vision encoder and open-source \llms{}. Recently, GPT-4V~\cite{gpt4-v} and Gemini~\cite{team2023gemini} were released, followed by work~\cite{yang2023dawn,zhang2023gpt4vision,tong2024eyes} that explore their impressive multimodal capabilities. LVIS-Instruct4V~\cite{wang2023see} and ShareGPT-4V~\cite{chen2023sharegpt4v} are created with GPT-4V for high-quality instruction tuning data. Our work implies the limitations of existing open-source \mllms{}, and provides a solution to enhance robust visual instruction following.

There is another thread of work~\cite{rust2022language,lee2023pix2struct,gao2024improving} emerged to explore the text along with images, charts, and tables all through visual input. Among of them, the most similar with ours is Pix2Struct~\cite{lee2023pix2struct}, however, there are a few difference: 1). Data-wisely, Pix2Struct takes the web-scale interleave data (HTML Dom Tree) for pretraining, we utilize the existing public image-text corpus (mainly GPT-4V synthetic corpus) for \short{} training. 2). Model-wisely, Pix2Struct employs an encoder-decoder model with BART-like learning signals, \vimmodel{} is an image encoder-decoder model (based on the pretrained \llm{} backbone) with autoregressive loss. 3). For downstream tasks, Pix2Struct does the finetuning before evaluation. While, \vimmodel{} training has no individual downstream finetuning.

\paragraph{Benchmarks} In parallel with the \mllms{} development, a trend emerges in proposing a variety of benchmarks. MME~\cite{fu2023mme} proposes 14 tasks with Yes/No questions based on the images. MMBench~\cite{liu2023mmbench} and SEED-Bench~\cite{li2023seed} introduce benchmarks that cover a variety of multiple-choice questions. MM-Vet~\cite{yu2023mm} extends to evaluate the open-ended outputs from \mllms{}. VisIT-Bench~\cite{bitton2023visit} accesses a range of tasks from recognition to complex reasoning, while Q-Bench~\cite{wu2023q} accesses low-level visual perception and understanding. MathVista~\citep{lu2023mathvista} focuses on systematically studying the mathematical reasoning capability in visual context. More recently, MMMU~\citep{yue2023mmmu} is proposed to evaluate multimodal models in a broad range of college-level subject knowledge. All these benchmarks focus on text instruction evaluation for \mllms{}, and our proposed \short{} integrates the text instruction into the visual modality space, asking for strong visual interpretation capability for embedded instruction recognition and following. Additionally, our \short{} is orthogonal with these existing benchmarks, can be seamlessly adapted to any of them.

\section{Conclusion}
In this work, we review the existing \mllms{} from a visual perspective, and present \short{}, a challenging setting to assess the visual instruction following ability of \mllmsfull{}. We adapt \short{} to eight benchmarks, leading to \short{}-Bench. Through in-depth probing under zero-shot setting, we observed a common issue for the existing open-source \mllms: all fall short in the \short{} setting, in most cases performing not as good as those in the original \tem{} setting. Furthermore, we train a \vimmodel, which demonstrates robust instruction following capabilities under text and visual modality instruction settings on all the tasks.

{
\small
\bibliographystyle{plain}
\bibliography{main}

\begin{thebibliography}{10}

\bibitem{awadalla2023openflamingo}
Anas Awadalla, Irena Gao, Josh Gardner, Jack Hessel, Yusuf Hanafy, Wanrong Zhu, Kalyani Marathe, Yonatan Bitton, Samir Gadre, Shiori Sagawa, et~al.
\newblock Openflamingo: An open-source framework for training large autoregressive vision-language models.
\newblock {\em arXiv preprint arXiv:2308.01390}, 2023.

\bibitem{bai2023qwenvl}
Jinze Bai, Shuai Bai, Shusheng Yang, Shijie Wang, Sinan Tan, Peng Wang, Junyang Lin, Chang Zhou, and Jingren Zhou.
\newblock Qwen-vl: A versatile vision-language model for understanding, localization, text reading, and beyond, 2023.

\bibitem{bigham2010vizwiz}
Jeffrey~P Bigham, Chandrika Jayant, Hanjie Ji, Greg Little, Andrew Miller, Robert~C Miller, Robin Miller, Aubrey Tatarowicz, Brandyn White, Samual White, et~al.
\newblock Vizwiz: nearly real-time answers to visual questions.
\newblock In {\em Proceedings of the 23nd annual ACM symposium on User interface software and technology}, pages 333--342, 2010.

\bibitem{bitton2023visit}
Yonatan Bitton, Hritik Bansal, Jack Hessel, Rulin Shao, Wanrong Zhu, Anas Awadalla, Josh Gardner, Rohan Taori, and Ludwig Schimdt.
\newblock Visit-bench: A benchmark for vision-language instruction following inspired by real-world use.
\newblock {\em arXiv preprint arXiv:2308.06595}, 2023.

\bibitem{chen2023shikra}
Keqin Chen, Zhao Zhang, Weili Zeng, Richong Zhang, Feng Zhu, and Rui Zhao.
\newblock Shikra: Unleashing multimodal llm's referential dialogue magic.
\newblock {\em arXiv preprint arXiv:2306.15195}, 2023.

\bibitem{chen2023sharegpt4v}
Lin Chen, Jisong Li, Xiaoyi Dong, Pan Zhang, Conghui He, Jiaqi Wang, Feng Zhao, and Dahua Lin.
\newblock Sharegpt4v: Improving large multi-modal models with better captions.
\newblock {\em arXiv preprint arXiv:2311.12793}, 2023.

\bibitem{vicuna2023}
Wei-Lin Chiang, Zhuohan Li, Zi~Lin, Ying Sheng, Zhanghao Wu, Hao Zhang, Lianmin Zheng, Siyuan Zhuang, Yonghao Zhuang, Joseph~E. Gonzalez, Ion Stoica, and Eric~P. Xing.
\newblock Vicuna: An open-source chatbot impressing gpt-4 with 90\%* chatgpt quality, March 2023.

\bibitem{chung2022scaling}
Hyung~Won Chung, Le~Hou, Shayne Longpre, Barret Zoph, Yi~Tay, William Fedus, Eric Li, Xuezhi Wang, Mostafa Dehghani, Siddhartha Brahma, et~al.
\newblock Scaling instruction-finetuned language models.
\newblock {\em arXiv preprint arXiv:2210.11416}, 2022.

\bibitem{dai2023instructblip}
Wenliang Dai, Junnan Li, Dongxu Li, Anthony Meng~Huat Tiong, Junqi Zhao, Weisheng Wang, Boyang Li, Pascale Fung, and Steven Hoi.
\newblock Instructblip: Towards general-purpose vision-language models with instruction tuning.
\newblock {\em arXiv preprint arXiv:2305.06500}, 2023.

\bibitem{fu2023mme}
Chaoyou Fu, Peixian Chen, Yunhang Shen, Yulei Qin, Mengdan Zhang, Xu~Lin, Zhenyu Qiu, Wei Lin, Jinrui Yang, Xiawu Zheng, et~al.
\newblock Mme: A comprehensive evaluation benchmark for multimodal large language models.
\newblock {\em arXiv preprint arXiv:2306.13394}, 2023.

\bibitem{gao2024improving}
Tianyu Gao, Zirui Wang, Adithya Bhaskar, and Danqi Chen.
\newblock Improving language understanding from screenshots.
\newblock {\em arXiv preprint arXiv:2402.14073}, 2024.

\bibitem{google2023bard}
Google.
\newblock Bard - chat based ai tool from google, powered by palm 2.
\newblock {\em https://bard.google.com/?hl=en}, 2023.

\bibitem{lee2023pix2struct}
Kenton Lee, Mandar Joshi, Iulia~Raluca Turc, Hexiang Hu, Fangyu Liu, Julian~Martin Eisenschlos, Urvashi Khandelwal, Peter Shaw, Ming-Wei Chang, and Kristina Toutanova.
\newblock Pix2struct: Screenshot parsing as pretraining for visual language understanding.
\newblock In {\em International Conference on Machine Learning}, pages 18893--18912. PMLR, 2023.

\bibitem{li2023seed}
Bohao Li, Rui Wang, Guangzhi Wang, Yuying Ge, Yixiao Ge, and Ying Shan.
\newblock Seed-bench: Benchmarking multimodal llms with generative comprehension.
\newblock {\em arXiv preprint arXiv:2307.16125}, 2023.

\bibitem{liu2023improved}
Haotian Liu, Chunyuan Li, Yuheng Li, and Yong~Jae Lee.
\newblock Improved baselines with visual instruction tuning.
\newblock {\em arXiv preprint arXiv:2310.03744}, 2023.

\bibitem{liu2023visual}
Haotian Liu, Chunyuan Li, Qingyang Wu, and Yong~Jae Lee.
\newblock Visual instruction tuning.
\newblock {\em arXiv preprint arXiv:2304.08485}, 2023.

\bibitem{liu2023mmbench}
Yuan Liu, Haodong Duan, Yuanhan Zhang, Bo~Li, Songyang Zhang, Wangbo Zhao, Yike Yuan, Jiaqi Wang, Conghui He, Ziwei Liu, et~al.
\newblock Mmbench: Is your multi-modal model an all-around player?
\newblock {\em arXiv preprint arXiv:2307.06281}, 2023.

\bibitem{lu2023mathvista}
Pan Lu, Hritik Bansal, Tony Xia, Jiacheng Liu, Chunyuan Li, Hannaneh Hajishirzi, Hao Cheng, Kai-Wei Chang, Michel Galley, and Jianfeng Gao.
\newblock Mathvista: Evaluating mathematical reasoning of foundation models in visual contexts.
\newblock {\em arXiv preprint arXiv:2310.02255}, 2023.

\bibitem{marino2019ok}
Kenneth Marino, Mohammad Rastegari, Ali Farhadi, and Roozbeh Mottaghi.
\newblock Ok-vqa: A visual question answering benchmark requiring external knowledge.
\newblock In {\em Proceedings of the IEEE/cvf conference on computer vision and pattern recognition}, pages 3195--3204, 2019.

\bibitem{masry2022chartqa}
Ahmed Masry, Do~Xuan Long, Jia~Qing Tan, Shafiq Joty, and Enamul Hoque.
\newblock Chartqa: A benchmark for question answering about charts with visual and logical reasoning.
\newblock {\em arXiv preprint arXiv:2203.10244}, 2022.

\bibitem{gpt4}
OpenAI.
\newblock Gpt-4: Technical report.
\newblock {\em arXiv preprint arXiv:2303.08774}, 2023.

\bibitem{gpt4-v}
OpenAI.
\newblock Gpt-4v(ision) system card.
\newblock {\em https://openai.com/research/gpt-4v-system-card}, 2023.

\bibitem{gpt-4o}
OpenAI.
\newblock Gpt-4o.
\newblock {\em https://openai.com/index/hello-gpt-4o}, 2024.

\bibitem{peng2023instruction}
Baolin Peng, Chunyuan Li, Pengcheng He, Michel Galley, and Jianfeng Gao.
\newblock Instruction tuning with gpt-4.
\newblock {\em arXiv preprint arXiv:2304.03277}, 2023.

\bibitem{rust2022language}
Phillip Rust, Jonas~F Lotz, Emanuele Bugliarello, Elizabeth Salesky, Miryam de~Lhoneux, and Desmond Elliott.
\newblock Language modelling with pixels.
\newblock {\em arXiv preprint arXiv:2207.06991}, 2022.

\bibitem{singh2019towards}
Amanpreet Singh, Vivek Natarajan, Meet Shah, Yu~Jiang, Xinlei Chen, Dhruv Batra, Devi Parikh, and Marcus Rohrbach.
\newblock Towards vqa models that can read.
\newblock In {\em Proceedings of the IEEE/CVF conference on computer vision and pattern recognition}, pages 8317--8326, 2019.

\bibitem{su2023pandagpt}
Yixuan Su, Tian Lan, Huayang Li, Jialu Xu, Yan Wang, and Deng Cai.
\newblock Pandagpt: One model to instruction-follow them all.
\newblock {\em arXiv preprint arXiv:2305.16355}, 2023.

\bibitem{sun2023aligning}
Zhiqing Sun, Sheng Shen, Shengcao Cao, Haotian Liu, Chunyuan Li, Yikang Shen, Chuang Gan, Liang-Yan Gui, Yu-Xiong Wang, Yiming Yang, et~al.
\newblock Aligning large multimodal models with factually augmented rlhf.
\newblock {\em arXiv preprint arXiv:2309.14525}, 2023.

\bibitem{taori2023stanford}
Rohan Taori, Ishaan Gulrajani, Tianyi Zhang, Yann Dubois, Xuechen Li, Carlos Guestrin, Percy Liang, and Tatsunori~B Hashimoto.
\newblock Stanford alpaca: An instruction-following llama model, 2023.

\bibitem{team2023gemini}
Gemini Team, Rohan Anil, Sebastian Borgeaud, Yonghui Wu, Jean-Baptiste Alayrac, Jiahui Yu, Radu Soricut, Johan Schalkwyk, Andrew~M Dai, Anja Hauth, et~al.
\newblock Gemini: a family of highly capable multimodal models.
\newblock {\em arXiv preprint arXiv:2312.11805}, 2023.

\bibitem{tong2024eyes}
Shengbang Tong, Zhuang Liu, Yuexiang Zhai, Yi~Ma, Yann LeCun, and Saining Xie.
\newblock Eyes wide shut? exploring the visual shortcomings of multimodal llms, 2024.

\bibitem{touvron2023llama}
Hugo Touvron, Thibaut Lavril, Gautier Izacard, Xavier Martinet, Marie-Anne Lachaux, Timoth{\'e}e Lacroix, Baptiste Rozi{\`e}re, Naman Goyal, Eric Hambro, Faisal Azhar, et~al.
\newblock Llama: Open and efficient foundation language models.
\newblock {\em arXiv preprint arXiv:2302.13971}, 2023.

\bibitem{touvron2023llama2}
Hugo Touvron, Louis Martin, Kevin Stone, Peter Albert, Amjad Almahairi, Yasmine Babaei, Nikolay Bashlykov, Soumya Batra, Prajjwal Bhargava, Shruti Bhosale, et~al.
\newblock Llama 2: Open foundation and fine-tuned chat models.
\newblock {\em arXiv preprint arXiv:2307.09288}, 2023.

\bibitem{wang2023see}
Junke Wang, Lingchen Meng, Zejia Weng, Bo~He, Zuxuan Wu, and Yu-Gang Jiang.
\newblock To see is to believe: Prompting gpt-4v for better visual instruction tuning.
\newblock {\em arXiv preprint arXiv:2311.07574}, 2023.

\bibitem{wu2023q}
Haoning Wu, Zicheng Zhang, Erli Zhang, Chaofeng Chen, Liang Liao, Annan Wang, Chunyi Li, Wenxiu Sun, Qiong Yan, Guangtao Zhai, et~al.
\newblock Q-bench: A benchmark for general-purpose foundation models on low-level vision.
\newblock {\em arXiv preprint arXiv:2309.14181}, 2023.

\bibitem{yang2023set}
Jianwei Yang, Hao Zhang, Feng Li, Xueyan Zou, Chunyuan Li, and Jianfeng Gao.
\newblock Set-of-mark prompting unleashes extraordinary visual grounding in gpt-4v.
\newblock {\em arXiv preprint arXiv:2310.11441}, 2023.

\bibitem{yang2023dawn}
Zhengyuan Yang, Linjie Li, Kevin Lin, Jianfeng Wang, Chung-Ching Lin, Zicheng Liu, and Lijuan Wang.
\newblock The dawn of lmms: Preliminary explorations with gpt-4v (ision).
\newblock {\em arXiv preprint arXiv:2309.17421}, 2023.

\bibitem{ye2023mplug}
Qinghao Ye, Haiyang Xu, Guohai Xu, Jiabo Ye, Ming Yan, Yiyang Zhou, Junyang Wang, Anwen Hu, Pengcheng Shi, Yaya Shi, et~al.
\newblock mplug-owl: Modularization empowers large language models with multimodality.
\newblock {\em arXiv preprint arXiv:2304.14178}, 2023.

\bibitem{you2023ferret}
Haoxuan You, Haotian Zhang, Zhe Gan, Xianzhi Du, Bowen Zhang, Zirui Wang, Liangliang Cao, Shih-Fu Chang, and Yinfei Yang.
\newblock Ferret: Refer and ground anything anywhere at any granularity.
\newblock {\em arXiv preprint arXiv:2310.07704}, 2023.

\bibitem{yu2023mm}
Weihao Yu, Zhengyuan Yang, Linjie Li, Jianfeng Wang, Kevin Lin, Zicheng Liu, Xinchao Wang, and Lijuan Wang.
\newblock Mm-vet: Evaluating large multimodal models for integrated capabilities.
\newblock {\em arXiv preprint arXiv:2308.02490}, 2023.

\bibitem{yue2023mmmu}
Xiang Yue, Yuansheng Ni, Kai Zhang, Tianyu Zheng, Ruoqi Liu, Ge~Zhang, Samuel Stevens, Dongfu Jiang, Weiming Ren, Yuxuan Sun, et~al.
\newblock Mmmu: A massive multi-discipline multimodal understanding and reasoning benchmark for expert agi.
\newblock {\em arXiv preprint arXiv:2311.16502}, 2023.

\bibitem{zhang2023llama}
Renrui Zhang, Jiaming Han, Aojun Zhou, Xiangfei Hu, Shilin Yan, Pan Lu, Hongsheng Li, Peng Gao, and Yu~Qiao.
\newblock Llama-adapter: Efficient fine-tuning of language models with zero-init attention.
\newblock {\em arXiv preprint arXiv:2303.16199}, 2023.

\bibitem{zhang2023gpt4vision}
Xinlu Zhang, Yujie Lu, Weizhi Wang, An~Yan, Jun Yan, Lianke Qin, Heng Wang, Xifeng Yan, William~Yang Wang, and Linda~Ruth Petzold.
\newblock Gpt-4v(ision) as a generalist evaluator for vision-language tasks, 2023.

\bibitem{zhu2023minigpt}
Deyao Zhu, Jun Chen, Xiaoqian Shen, Xiang Li, and Mohamed Elhoseiny.
\newblock Minigpt-4: Enhancing vision-language understanding with advanced large language models.
\newblock {\em arXiv preprint arXiv:2304.10592}, 2023.

\end{thebibliography}
}


\appendix
\clearpage
\addcontentsline{toc}{section}{Appendix} 
\part{Appendix} 
\parttoc 
\clearpage

\renewcommand{\contentsname}{Supplemental Materials}

\section{Benchmark Details}
\label{sec:benchmark_details}

\subsection{Source Datasets}
\label{sec:app_source_datasets}
We consider eight representative datasets, MM-Vet~\cite{yu2023mm},  MME~\cite{fu2023mme}, OKVQA, VizWiz, TextVQA, MathVista, ChartQA, and MMMU. We also provide probing analysis on VQA$_{v2}$ \textit{test-dev} split, RefCOCO \textit{testA} split, RefCOCO+ \textit{testA} split, and RefCOCOg \textit{test} split to better illustrate how the data is formatted.

\begin{figure}[ht]
\centering
\includegraphics[width=0.8\textwidth]{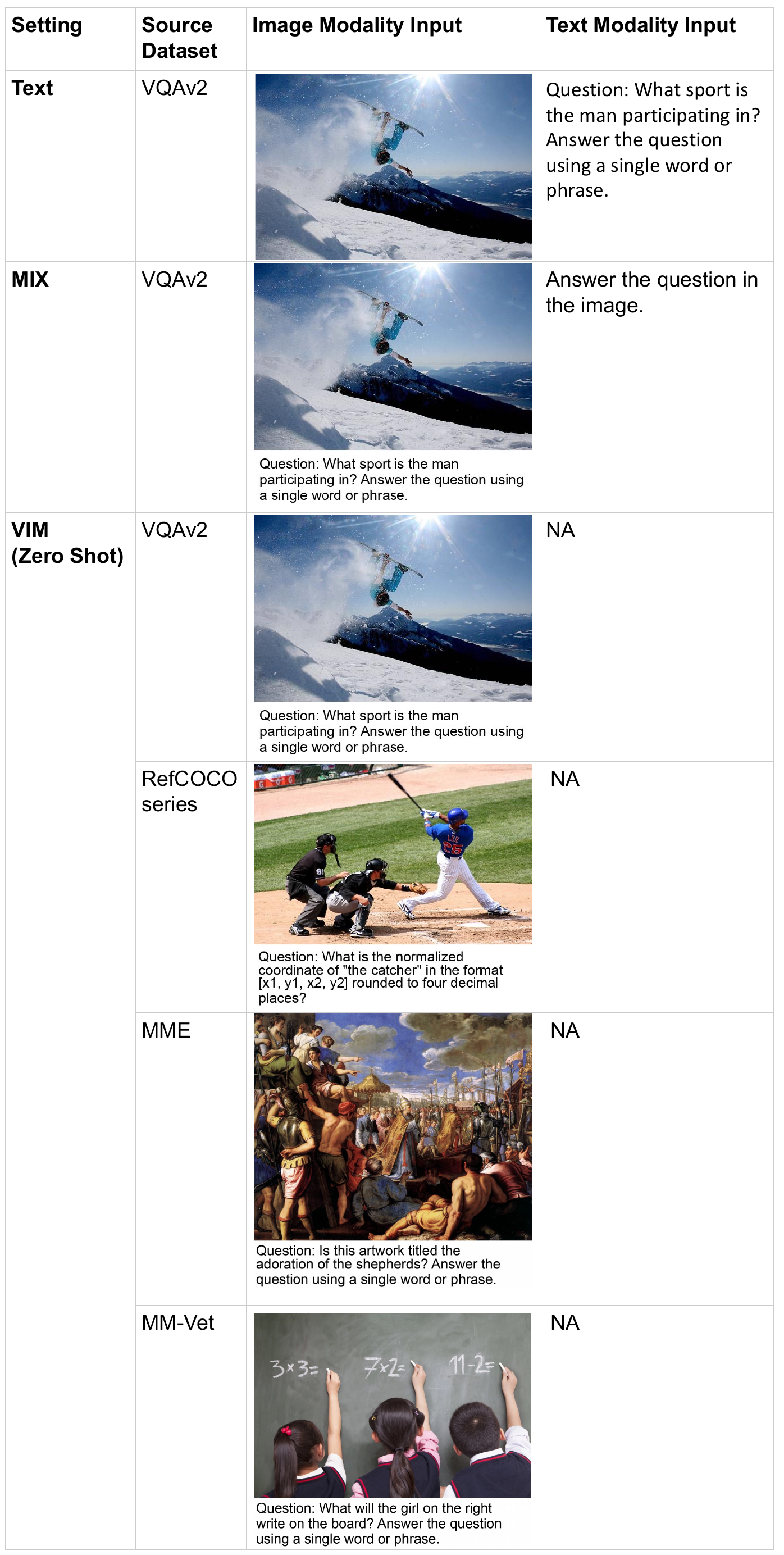}
\vspace{-3mm}
\caption{Dataset example comparison of three instruction probing settings: Text, MIX and \short{}.
}
\label{fig:app_source_dataset_example}
\end{figure}

In Figure~\ref{fig:app_source_dataset_example}, we showcase an example from the source dataset VQA$_{v2}$, which is with instruction probing setting ``Text''. We also consider other two probing settings, ``MIX'' that have an additional text prompt to guide the visual embedded instruction following, and our proposed ``\short{}'' that only allow the image with embedded instruction as input.
We also showcases dataset examples sampled from source datasets RefCOCO, MME, and MM-Vet. All the \short{} test samples does not include any additional instruction input in text modality (noted as ``NA'').

\subsection{Dataset Processing Pipeline}
\label{sec:app_dataset_processing}

For each source dataset, we start by building up zero shot by embedding instructions into the input image to concatenate as a new image which contains instructions in image modality. In this way, we obtain a new image with embedded instructions for each image-question pair.


\subsection{Evaluation Details}
\label{sec:app_evaluation_detail}

\paragraph{Metrics}
We follow the evaluation pipeline for each benchmark. We use parsing and accuracy for MathVista and MMMU with a more strict protocol. For example, when the model is outputting an empty or random string for a multiple-choice question, in MathVista, the original evaluation protocol will use Levenshtein distance to map to a most similar prediction option, and in MMMU, a random choice from the candidate list will be applied. This will introduce noise and randomness for the evaluation, may not correctly reflect the model performance. In our strict evaluation protocol, we eliminate this random match strategy. 

For OKVQA and TextVQA, we follow the leaderboard evaluation to use an evaluation metric that is robust to inter-human variability: $\text{Acc}(\text{ans}) = \min \left\{ \frac{\#\text{humans that said ans}}{3}, 1 \right\}$. For ChartQA, we use relaxed accuracy on human and augmented split.

For MME, the standard metric ($Score$) proposed in ~\cite{fu2023mme} is the summed up Accuracy ($Acc$) and Accuracy+ ($Acc_{+}$) as: $Score = sum(Acc \times 100\%, Acc_{+} \times 100\%)$, where the former one count each correct answer as correct, while the latter one only considers correct when both ``Yes'' and ``No'' questions for each image are answered correctly.

For MM-Vet, we use GPT-4 (``gpt-4-0613'' version) to automatically provide the score for each sample. The final Accuracy reported as: $\text{Acc} = \frac{\sum_{i=1}^{N} s_i}{N} \times 100\%
$, where $s_i$ is the score at scale $0-1$ for sample $i$.


\begin{table}[ht!]
\caption{Zero shot evaluation results of \texttt{Text} probing setting on VQA$_{v2}$, MME, MM-vet. This is the popular setting for evaluating text instruction following capability of \mllms{}, where the input image and text instruction are both provided.
}
\label{tab:app_image_w_instruction}
    \centering
    \resizebox{\columnwidth}{!}{
\begin{tabular} 
{l@{\hspace{2mm}}cc|c@{\hspace{2mm}}c@{\hspace{2mm}}c@{\hspace{2mm}}c}
\toprule
\multirow{2}{*}{Models} & \multirow{2}{*}{LLM} & Embedded & \multicolumn{3}{c}{\textit{Zero shot}} \\
 & & Instruction & VQA$_{v2}$ & MME & MM-Vet  \\
\hline
\rowcolor{magnolia} \multicolumn{6}{c}{\texttt{Sub set}} \\
\multirow{4}{*}{LLaVA-1.5} & {\multirow{2}{*}{\footnotesize Vicuna-7b}} & w.o. & $60.75$ & $108$  & $31.3$  \\ 
&  & w. & $57.88$ & $88$  & $27.9$  \\
\cmidrule{2-6}
& {\multirow{2}{*}{\footnotesize Vicuna-13b}} & w.o. & $61.00$ & $106$ & $35.2$  \\
&  & w. & $58.00$ & $87$ & $32.7$  \\
\bottomrule
\end{tabular}
    }
\end{table}

\section{Full Analysis}
\subsection{Robustness Analysis}
\subsubsection{Prompt for \texttt{MIX} Probing Setting}

In \texttt{mix} probing setting, the \mllms{} can accept an extra text instruction input as guidance. The model performance may vary when given different prompts. We report the results using four relevant but diversified prompts (Prompt \#1-\#5) in Table~\ref{tab:mix_prompt_robustness}. To be specific, the detailed prompts we use are: 1) Prompt \#1: \textit{``Answer the question in the image.''}, 2) Prompt \#2: \textit{``Please answer the question that is written in the image.''}, 3) Prompt \#3: \textit{``Follow the instruction embedded in the image.''}, 4) Prompt \#4: \textit{``Detect the question in the image and directly answer to it.''}.


\subsubsection{Image Embedded with Instruction}
To investigate whether the model performance is robust to the minimal changes introduced by the embedded instruction, we give both the original instruction in the text modality and the image with instruction embedded as the image modality to the model.

In Table~\ref{tab:app_image_w_instruction}, we present comparative results of LLaVA-1.5 using Vicuna-7b and Vicuna-13b language backbones. It's observed that embedding images with instructions leads to a marginal decline in performance. This trend suggests that current \mllms{} may not be entirely robust to variations in images. However, this performance degradation is minor and within acceptable limits. This implies that the disparity in performance between the \texttt{Text} and \texttt{VIM} probing settings is not solely attributable to changes in the images, but is largely due to the models' capacity to follow visual embedded instructions.

\begin{table}[!ht]
\caption{One ``failure" examples to show the instruction recognition of GPT-4V.}
\label{tab:ab_instr_rec}
\centering
\begin{tcolorbox}[colback=splashedwhite,boxsep=1pt,left=2pt,right=0pt,top=5pt,bottom=3pt,boxrule=0.5pt] 
\centering
\resizebox{\textwidth}{!}{
\begin{tabular}{p{1.0\columnwidth} c}
\VarSty{ {\bf \textit{Image w. Embedded Instruction}} } & \\
\multirow{4}{*}{
\includegraphics[width=1.0\textwidth]{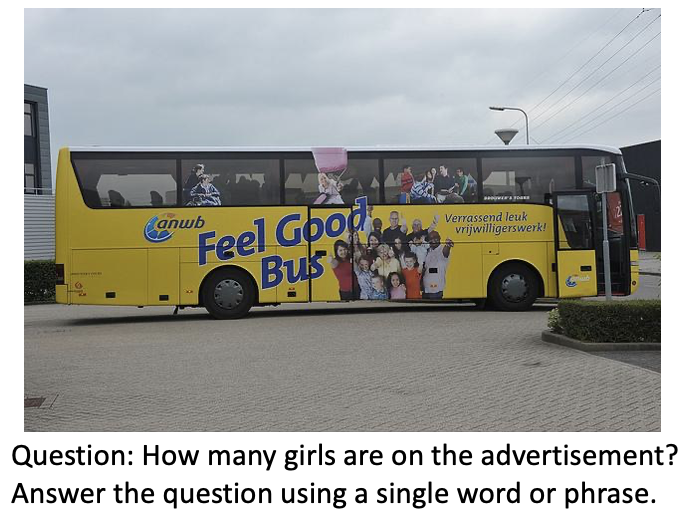} 
} \\
\\
\vspace{9.3cm}
\VarSty{ {\bf Text Prompt}:} & \\
What is the text in the image? \\
\vspace{1mm}
\VarSty{ {\bf GPT-4V}:}  & \\
1. "anwb" \\
2. "Feel Good Bus" \\
3. "Verrassend leuk vrijwilligerswerk!" \\
\end{tabular}
}
\end{tcolorbox}
\end{table}

\begin{table}[!ht]
\caption{Carefully curated prompting to unleash the grounding capability of GPT-4V.}
\label{tab:dis_gpt4}
\centering
\begin{tcolorbox}[colback=splashedwhite,boxsep=1pt,left=2pt,right=2pt,top=5pt,bottom=3pt,boxrule=1pt] 
\centering
\resizebox{\textwidth}{!}{
\begin{tabular}{p{1.0\columnwidth} c}
\VarSty{ {\bf \textit{Image w.o. Embedded Instruction}} } & \\
\multirow{4}{*}{
\includegraphics[width=1.0\textwidth]{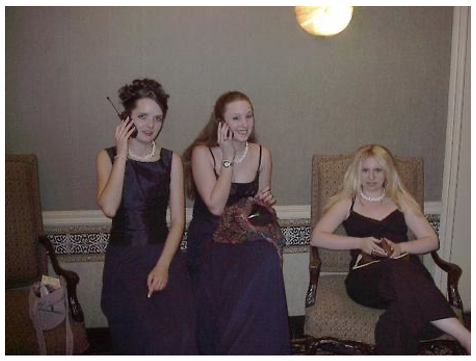} 
} \\
\\
\vspace{9.3cm}
\VarSty{ {\bf Text Prompt}:} & \\
How many people in the image? Give them a label and show me their normalized coordinate in the format [x1, y1, x2, y2] rounded to four decimal places. \\
\vspace{1mm}
\VarSty{ {\bf GPT-4V}:}  & \\
There are 3 people in the image. \\
Labels and normalized coordinates for the people are: \\
1. Person A: [0.1081, 0.0432, 0.3784, 0.7973] \\
2. Person B: [0.3838, 0.0000, 0.7027, 0.8108] \\
3. Person C: [0.7027, 0.2973, 0.9676, 0.8108] \\
\end{tabular}
}
\end{tcolorbox}
\end{table}

\begin{table*}[ht!]
\caption{Zero shot evaluation results on MM-Vet with detail scores across six core vision and language capabilities. \colorbox{Gray}{\makebox(5,5){}} highlights the \textcolor{dimgray}{numbers} for \texttt{mix} instruction setting. The numbers are presented in $\%$. Rec: Recognition, Know: Knowledge, Gen: Language Generation, Spat: Spatial Awareness.
}
\label{tab:app_mmvet_details}
    \centering
    \resizebox{.8\textwidth}{!}{
    \begin{tabular}{l|cccccc|c}
    \hline
        ~ & Rec & OCR & Know & Gen & Spat & Math & Total \\ \hline
        LLaVA-1.5-7b & $12$ & $6.6$ & $5.6$ & $5.6$ & $6.4$ & $3.8$ & $10.1$ \\
        \rowcolor{Gray} & \textcolor{dimgray}{$9.7$} & \textcolor{dimgray}{$7.7$} & \textcolor{dimgray}{$5.1$} & \textcolor{dimgray}{$3.1$} & \textcolor{dimgray}{$6.9$} & \textcolor{dimgray}{$3.8$} & \textcolor{dimgray}{$8.5$} \\ \hline
        LLaVA-1.5-13b & $15.2$ & $13.6$ & $6.9$ & $10.9$ & $11.9$ & $3.8$ & $14.4$ \\
        \rowcolor{Gray} & \textcolor{dimgray}{$18.5$} & \textcolor{dimgray}{$15.7$} & \textcolor{dimgray}{$7.3$} & \textcolor{dimgray}{$9.5$} & \textcolor{dimgray}{$15.3$} & \textcolor{dimgray}{$9.6$} & \textcolor{dimgray}{$16.9$} \\ \hline
        InstructBLIP & $6.4$ & $1.6$ & $1.8$ & $1.2$ & $2.7$ & $0$ & $4.4$ \\ 
        \rowcolor{Gray} & \textcolor{dimgray}{$14.5$} & \textcolor{dimgray}{$7.8$} & \textcolor{dimgray}{$2.6$} & \textcolor{dimgray}{$0.9$} & \textcolor{dimgray}{$9.3$} & \textcolor{dimgray}{$11.5$} & \textcolor{dimgray}{$12.5$} \\ \hline
        GPT-4V & $61.4$ & $65.2$ & $51.2$ & $53.7$ & $67.6$ & $59.2$ & $63.5$ \\ \hline
    \end{tabular}
    }
\end{table*}

\begin{table*}[ht!]
\caption{Zero shot evaluation results on MME subset under \texttt{mix} setting. We compare the performance of LLaVA-1.5-7b and LLaVA-1.5-13b across four different prompts.}
    \label{tab:mix_prompt_robustness}
    \centering
    \resizebox{1.0\textwidth}{!}{
    \begin{tabular}{l|cccc|cccc}
    \toprule
        \multirow{2}{*}{Task} & \multicolumn{4}{c}{LLaVA-1.5-7b} & \multicolumn{4}{c}{LLaVA-1.5-13b} \\
        \cmidrule{2-9}
         & Prompt \#1 & Prompt \#2 & Prompt \#3 & Prompt \#4 & Prompt \#1 & Prompt \#2 & Prompt \#3 & Prompt \#4 \\ \hline
        artwork & 5 & 5 & 0 & 0 & 0 & 0 & 0 & 0 \\
        celebrity & 5 & 5 & 0 & 0 & 1 & 0 & 0 & 0 \\
        code reasoning & 4 & 3 & 0 & 0 & 0 & 0 & 0 & 0 \\
        color & 5 & 5 & 0 & 0 & 1 & 1 & 0 & 0 \\
        commonsense reasoning & 4 & 4 & 0 & 0 & 0 & 0 & 0 & 0 \\
        count & 5 & 5 & 0 & 0 & 1 & 0 & 0 & 0 \\
        existence & 6 & 6 & 0 & 0 & 0 & 0 & 0 & 0 \\
        landmark & 5 & 5 & 0 & 0 & 3 & 0 & 0 & 0 \\
        numerical calculation & 5 & 5 & 0 & 0 & 0 & 0 & 0 & 0 \\
        OCR & 5 & 5 & 0 & 0 & 0 & 0 & 0 & 0 \\
        position & 4 & 5 & 0 & 0 & 0 & 0 & 0 & 0 \\
        posters & 5 & 5 & 0 & 0 & 1 & 0 & 0 & 0 \\
        scene & 5 & 5 & 0 & 0 & 0 & 0 & 0 & 0 \\
        text translation & 5 & 5 & 0 & 0 & 0 & 0 & 0 & 0 \\ \hline
        $Correct$ & 68 & 68 & 0 & 0 & 7 & 1 & 0 & 0 \\
    \bottomrule
    \end{tabular}
    }
\end{table*}

\begin{table}[t!]
\caption{\mllms{}'s response to the question \#202 in MM-Vet.}
\label{tab:ab_people_q}
\centering
\begin{tcolorbox}[colback=splashedwhite,boxsep=1pt,left=2pt,right=15pt,top=5pt,bottom=3pt,boxrule=1pt] 
\centering
\small
\begin{tabular}{p{1.0\columnwidth} c}
\VarSty{ {\bf {\textit{Image w. Embedded Instruction}}} } & \\ 
\\
\vspace{-6mm}
\multirow{4}{*}{
\includegraphics[width=.8\textwidth]{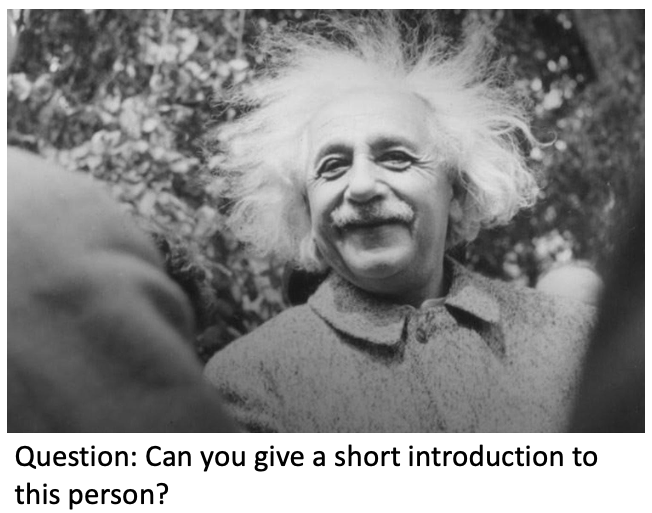}
} \\
\\
\vspace{7.2cm}
\VarSty{ {\bf GPT-4V:}} Sorry, I cannot help with that. \\
\VarSty{ {\bf LLaVA-1.5:}} Question: Can you give a short introduction to this person?\\
\VarSty{ {\bf InstructBLIP:}} albert einstein with the caption can you give a short introduction to this person? \\
\VarSty{ {\bf \vimmodel:}} albert einstein \\
\end{tabular}
\end{tcolorbox}
\end{table}

\subsection{Details of Instruction Recognition}
\label{sec:app_inst_rec}
Based on Section~\ref{sec:instr_rec}, we conduct an ablation to verify the instruction recognition capability of these \mllms{}. Table~\ref{tab:app_zs_instr_rec} showcases some example results for zero shot instruction recognition. GPT-4V can recognize the text instruction in both settings, LLaVA can detect some words of the instructions, but may not perfectly recognize the instructions, especially in the one shot setting. Table~\ref{tab:ab_instr_rec} shows a failure example of GPT-4V for Instruction Recognition, it recognizes the logo texts on the bus as the text instruction.

\begin{table*}[t!]
\caption{Zero Shot Instruct Recognition: \mllms{}'s recognition to the example questions in VQA.}
\label{tab:app_zs_instr_rec}
\centering
\begin{tcolorbox}[colback=splashedwhite,boxsep=1pt,left=2pt,right=5pt,top=5pt,bottom=3pt,boxrule=0.5pt] 
\centering
\small
\resizebox{\textwidth}{!}{
\begin{tabular}{p{0.5\textwidth} p{0.5\textwidth} } 
\VarSty{ {\bf \textit{Image w. Embedded Instruction}} } {\textit{\#42000} in VQA} &  \VarSty{ {\bf \textit{Image w. Embedded Instruction}} }  {\textit{\#757000} in VQA}\\
\multirow{4}{*}{
\includegraphics[width=0.5\textwidth]{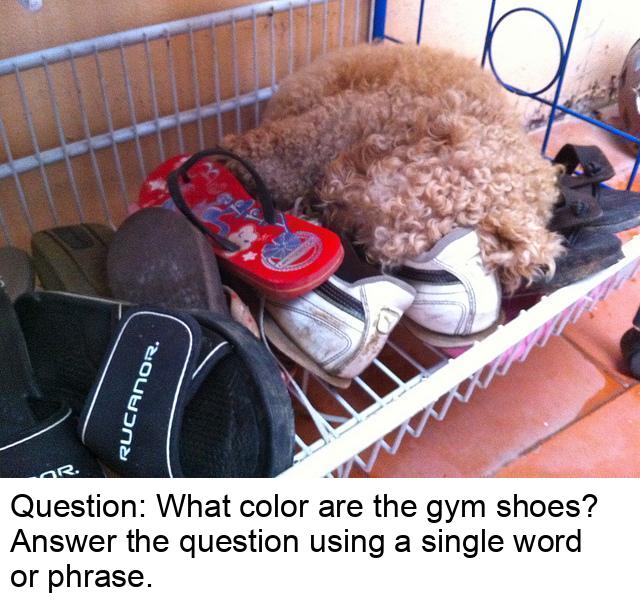}
} & 
\multirow{4}{*}{
\includegraphics[width=0.5\textwidth]{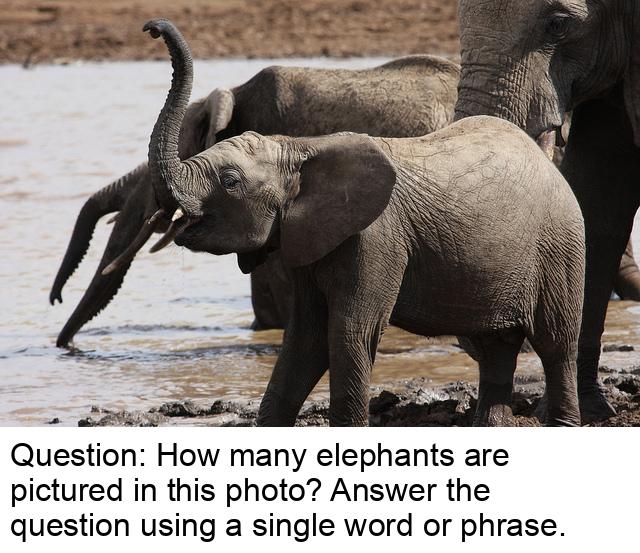}
} \\
\rule{0pt}{6.0cm} & \\
\VarSty{ {\bf Text Prompt:}} What is the text in the image? & 
\VarSty{ {\bf Text Prompt:}} What is the text in the image? \\ 
\\

\VarSty{ {\bf GPT-4V:}} Question: What color are they gym shoes? Answer the question using a single word or phrase.  & 
\VarSty{ {\bf GPT-4V:}} Question: How many elephants are pictured in this photo? Answer the question using a single word or phrase. \\
\VarSty{ {\bf LLaVA-1.5-7B:}} The text in the image is a question asking, ``What color are the gym shoes?"  & 
\VarSty{ {\bf LLaVA-1.5-7B:}} The text in the image is a question asking how many elephants are pictured in the photo. \\
\VarSty{ {\bf LLaVA-1.5-13B:}} The text in the image is a question asking about the color of the gym shoes. & 
\VarSty{ {\bf LLaVA-1.5-13B:}} The text in the image is a question asking how many elephants are pictured in the photo. \\

\rule{0pt}{8mm} & \\

\VarSty{{\bf \textit{Image w. Embedded Instruction}}} {\textit{\#132132000} in VQA} &  \VarSty{{\bf \textit{Image w. Embedded Instruction}}} {\textit{\#131612000} in VQA} \\
\multirow{4}{*}{
\includegraphics[width=0.475\textwidth]{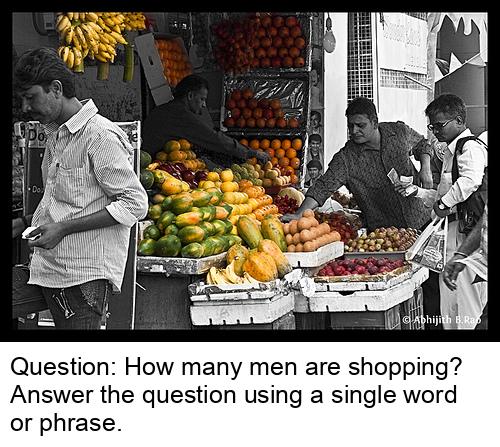}
} & 
\multirow{4}{*}{
\includegraphics[width=0.475\textwidth]{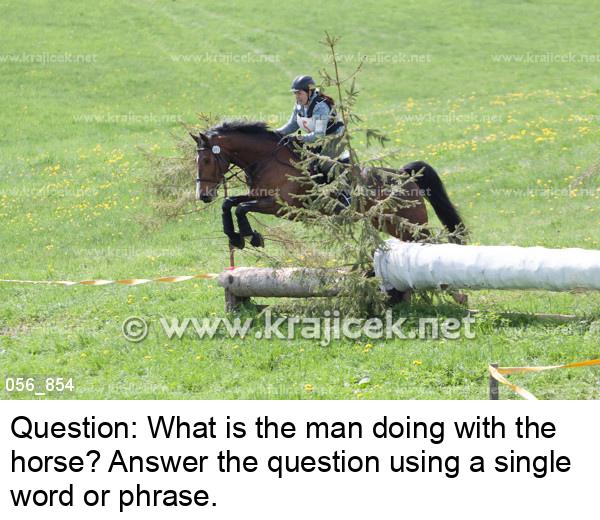}
} \\
\rule{0pt}{5.3cm} & \\
\VarSty{ {\bf Text Prompt:}} What is the text in the image? & 
\VarSty{ {\bf Text Prompt:}} What is the text in the image? \\ 
\\

\VarSty{ {\bf GPT-4V:}} Question: How many men are shopping? Answer the question using a single word or phrase.  & 
\VarSty{ {\bf GPT-4V:}} Question: What is the man doing with the horse? Answer the question using a single word or phrase. \\
\VarSty{ {\bf LLaVA-1.5-7B:}} The text in the image is a question asking, ``How many men are shopping?"  & 
\VarSty{ {\bf LLaVA-1.5-7B:}} The text in the image is a question asking, ``What is the man doing with the horse?" \\
\VarSty{ {\bf LLaVA-1.5-13B:}} The text in the image is a question asking how many men are shopping. & 
\VarSty{ {\bf LLaVA-1.5-13B:}} The text in the image is a caption that reads ``Question: What is the man doing with the horse? Answer the question using a single word or phrase." \\

\end{tabular}
}
\end{tcolorbox}
\end{table*}

\subsection{Qualitative Observations}
\label{sec:app_qual_ob}

\subsubsection{People related questions} In VQA and MM-Vet, there are some categories of questions about people or movies. For example, in Table~\ref{tab:ab_people_q}, GPT-4V will response ``\textit{Sorry, I cannot help with that.}'', while, LLaVA-1.5 and InstructBLIP just simply repeat the question. 

\subsubsection{Grounding of GPT-4V}
\label{sec:app_grounding_gpt4}

Table~\ref{tab:dis_gpt4} shows that GPT-4V's grounding capability can be unleashed when carefully prompted.

\section{Limitations}
\label{sec:limit}
We discuss the limitations of our work as follows: 1). Though \vimmodel{} exhibits robust instruction following capability in both the \tem{} and \short{} settings, it still has a gap with proprietary models, especially GPT-4V, there is still space to improve to be a generalist model. 2). In this work, the evaluation protocols and metrics of \short{} setting follow these from the original \tem{} setting, it also inherits the shortcoming of these evaluations, we leave these for future work. 3). For proprietary models (GPT-4V and Gemini Pro), they are evolving with in-context learning as more queries are fed to the models, the results from these models may not be stable or potentially reproducible.

\section{Broader Impacts}
\label{sec:impact}

\paragraph{Evaluation and Safety}
Safety is a critical aspect of \mllms, particularly when these models are used in the real-world applications. The safety of \mllms{} is assessed through various benchmarks that test their robustness against unsafe instructions and harmful content. \short{} identified a common issue for the existing open-source \mllms, it may help to improve the robustness of the \mllms{}; and also improve the current evaluation benchmarks of \mllms.

\paragraph{Ethical Considerations}
The deployment of \mllms{} necessitates careful consideration of ethical implications, including privacy, bias, and the potential misuse of technology. Ensuring that these models are properly developed and used responsibly is crucial to mitigate risks and maximize their positive impact on society.

\paragraph{Practical Applications}
\mllms{} have a wide range of applications, from enhancing accessibility in technology to improving human-computer interactions. There are many real application scenarios for \short{}, like UI interface navigation, agent development etc. This can improve efficiency, accuracy, and safety in manufacturing, logistics, and other sectors where visual instruction is crucial. We hope the \short{} can benefit for the MLLM's evolution.

\end{document}